\documentclass[10pt,journal,compsoc]{IEEEtran}

\ifCLASSOPTIONcompsoc
  \usepackage[nocompress]{cite}
\else
  \usepackage{cite}
\fi

\usepackage{makecell}
\newcommand{\PreserveBackslash}[1]{\let\temp=\\#1\let\\=\temp}
\newcolumntype{C}[1]{>{\PreserveBackslash\centering}p{#1}}
\newcolumntype{R}[1]{>{\PreserveBackslash\raggedleft}p{#1}}
\newcolumntype{L}[1]{>{\PreserveBackslash\raggedright}p{#1}}

\usepackage{graphicx}
\usepackage{amsmath,amssymb} 
\usepackage{multirow}
\usepackage{array}
\usepackage{bbding}

\usepackage{times}
\usepackage{epsfig}
\usepackage{mathrsfs}
\usepackage{multirow}
\usepackage[labelsep=period]{caption}
\usepackage{subfig}
\usepackage{url}
\usepackage{cite}

\ifCLASSINFOpdf
  
\else
 
\fi
\begin{document}

\title{Feature Completion for Occluded Person Re-Identification}

\author{Ruibing~Hou,~\IEEEmembership{Student Member,~IEEE},
        Bingpeng~Ma, Hong~Chang,~\IEEEmembership{Member,~IEEE},
        Xinqian~Gu,~\IEEEmembership{Student Member,~IEEE}, 
        Shiguang~Shan,~\IEEEmembership{Senior~Member,~IEEE}, 
        and~Xilin~Chen,~\IEEEmembership{Fellow,~IEEE}
\IEEEcompsocitemizethanks{\IEEEcompsocthanksitem R. Hou, H. Chang, X. Gu, S. Shan and X. Chen are with Key Lab of Intelligent Information Processing of Chinese Academy of Sciences (CAS), Institute of Computing Technology, CAS, Beijing, 100190, China and University of Chinese Academy of Sciences, Beijing 100049, China.

\IEEEcompsocthanksitem B. Ma is with the School of Computer Science and Technology, University of Chinese Academy of Sciences, Beijing 100049, China (e-mail: bpma@ucas.ac.cn).

\IEEEcompsocthanksitem S. Shan is also with the CAS Center for Excellence in Brain Science and Intelligence Technology, Shanghai, 200031, China. \protect\\
E-mail: \{ruibing.hou, xinqian.gu\}@vipl.ict.ac.cn, bpma@ucas.ac.cn, \{changhong, sgshan, xlchen\}@ict.ac.cn}}

\markboth{Journal of \LaTeX\ Class Files,~Vol.~14, No.~8, August~2015}%
{Shell \MakeLowercase{\textit{et al.}}: Bare Demo of IEEEtran.cls for Computer Society Journals}

\IEEEtitleabstractindextext{%
\begin{abstract}
Person re-identification (reID) plays an important role in computer vision. However, existing methods suffer from performance degradation in occluded scenes. In this work, we propose an  occlusion-robust block, \textit{Region Feature Completion} (RFC), for occluded reID. Different from most previous works that discard the occluded regions, RFC block can recover the semantics of occluded regions in feature space. Firstly, a \textit{Spatial RFC} (SRFC) module is developed.  SRFC exploits the long-range spatial contexts from non-occluded regions to predict the features of occluded regions.  The unit-wise prediction task leads to an encoder/decoder architecture, where the region-encoder models the correlation between non-occluded and occluded region, and the region-decoder utilizes the spatial correlation to recover occluded region features. Secondly, we introduce \textit{Temporal RFC} (TRFC) module which captures the long-term temporal contexts to refine the prediction of SRFC. RFC block is lightweight, end-to-end trainable and can be easily plugged into existing CNNs to form RFCnet. Extensive experiments are conducted on occluded and commonly holistic reID benchmarks. Our method significantly outperforms existing methods on the occlusion datasets, while remains top even superior performance on holistic datasets. The source code is available at \url{https://github.com/blue-blue272/OccludedReID-RFCnet}.
\end{abstract}

}

\maketitle

\IEEEdisplaynontitleabstractindextext

\IEEEpeerreviewmaketitle

\IEEEraisesectionheading{\section{Introduction}\label{sec:introduction}}
Person re-identification (reID) aims at re-identifying a target person across multiple non-overlapped cameras. This task has drawn increasing attention in recent years due to its importance in application, such as video surveillance. It remains a challenging problem because of complex variations in camera viewpoints, human poses, background clutter and occlusion.

Most of existing methods focus on holistic (non-occluded) images, while neglecting occluded images. In a real scenario, persons can be frequently occluded by some obstacles, \textit{e.g.} vehicles, trees or other persons, leading to occluded targets. Thus it is necessary to identify persons with occlusions, which is known as occluded person reID~\cite{zhuo2018occluded,miao2019pose}.

Compared to match persons with holistic images, occluded person reID is more challenging due to the interference of obstacles and information loss of the target person. Recently, some occluded person reID methods are proposed. Most of them discard the occluded parts. For example, the methods~\cite{sun2019perceive,zhuo2018occluded,he2019foreground}  detect non-occluded body parts using body part detectors and then only consider the shared visible parts for matching. However, these non-occluded parts based methods degrade reliability of the retrieval results. Therefore, these methods may fail in the situation, where the non-occluded parts share a similar appearance and the occluded body parts are key discriminative facts.

Another type of methods attempts to recover the appearance of the occluded parts. Our previous work~\cite{VRSTC}, VRSTC, employs an image completion network to predict the appearance of the occluded regions. With the completion, VRSTC can obtain an integral representation of the target person. The completion-based method provides more complete spatial information and coherent temporal information which benefits reID tasks.

Despite the significant performance improvement by the completion in occluded reID, VRSTC has some weaknesses due to image level completion. Although image completion is superior in interpreting and visualizing what is recovered, the feature level completion is more effective and efficient for improving reID performance. \textbf{Firstly, feature completion can be seamlessly integrated to the reID network whereas image completion cannot}. Because of the huge network parameters, the image completion is usually post-hoc addition to the reID network. It leads to that the reID task cannot provide direct feedback to the completion task. Therefore, the image completion is more dedicated to generating real images rather than improving the reID performance. In contrast, the feature completion is lightweight. It can be inserted into the reID backbone at any depth, which enables end-to-end training. In this way, the feedback signals from the reID task can supervise the feature completion to fully improve the reID performance.

\begin{figure}[t]
\centering
   \includegraphics[width=1.0\linewidth]{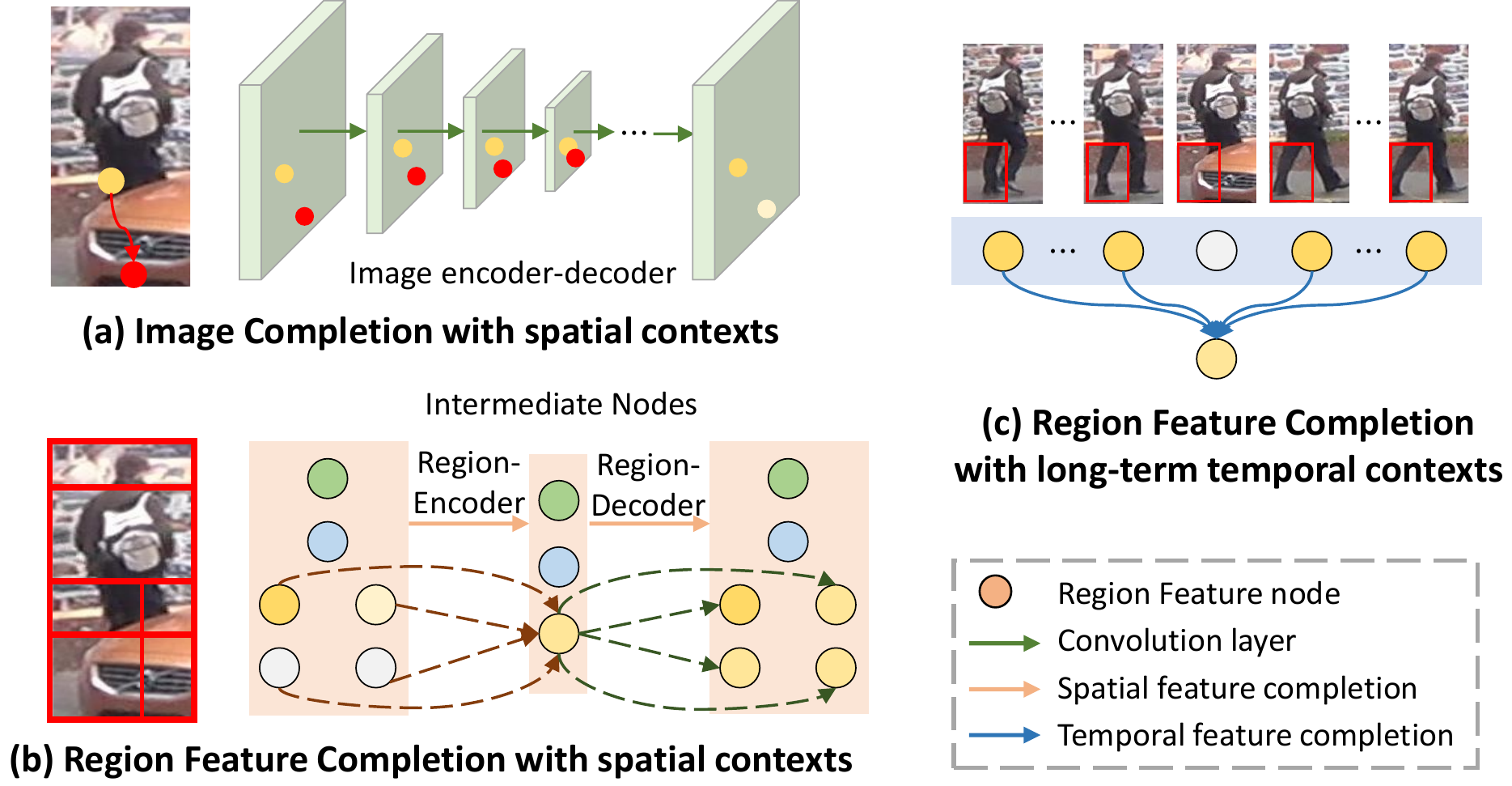}
   \caption{The motivation of our work. (a) Image completion network. It is built on an encoder-decoder pipeline. The information of the visible ``yellow'' pixel  can only be propagated to the occluded ``red'' pixel by \textit{stacking  multiple convolutional layers}, which is computationally inefficient. (b) Region feature completion with spatial contexts. By representing each body region as a feature, one \textit{single} spatial completion module can propagate the information among any two positions. (c) Region feature completion with long-term temporal contexts.}
\label{motivation}
\end{figure}

\textbf{Secondly, feature completion is more effective to capture long-range spatial contexts and long-term temporal contexts.} In general, image completion network is built on U-Net~\cite{context} that is an encoder-decoder pipeline. The encoder and decoder both consist of a series of convolutional layers. \textit{\textbf{For one thing}}, since the convolutional operations process a local spatial neighborhood, the long-range spatial contexts can only be captured when the convolutional operations are applied repeatedly. For example, as shown in Fig.~\ref{motivation} (a), the information of  the visible ``yellow'' pixel  can only be propagated to the occluded ``red'' pixel by stacking  multiple convolutional layers.  However, Wang \textsl{et al.}~\cite{non-local} point out that repeating convolution operations is computationally inefficient and causes optimization difficulties. On the contrary, as shown in Fig.~\ref{motivation} (b), the image feature map can be represented by a few region features through  \textit{region division}.  With a few feature nodes, it is easier for one feature completion module to propagate the spatial information among any two positions. Thus the feature completion can capture the distant spatial contexts more effectively. 

\textit{\textbf{For another thing}}, due to the limitation of computation resources, the expensive image completion operation usually only considers adjacent two frames. Instead, as shown in Fig.~\ref{motivation} (c), the feature completion only requires to predict a few region features, which imposes slight computations. So feature completion can use more adjacent frames, which captures longer-term temporal contexts to perform better completion.  In summary, it is appealing to explore a way of recovering occluded regions in feature level for occluded reID problem.

To this end, we propose a feature completion block, \textit{Region Feature Completion} (RFC), for video-based occluded person reID. \textbf{Firstly}, we design an \textit{Adaptive Partition Unit} (APU) to adaptively divide input feature maps into different regions. Each region corresponds to a specific body part, which can achieve precise spatial alignment. \textbf{Secondly},  we apply a \textit{Spatial Region Feature Completion} (SRFC) module, which uses spatial information to recover the occluded regions via a region-encoder and a region-decoder. As shown in Fig.~\ref{motivation} (b), the region-encoder learns to aggregate the occluded region and correlated non-occluded regions to an intermediate node. Then the region-decoder predicts the feature of the occluded region from this intermediate node. With these intermediate nodes, the information from correlated non-occluded regions can be propagated to the occluded region to recover its feature representation. \textbf{Thirdly}, we utilize a \textit{Temporal Region Feature Completion} (TRFC) module, which uses long-term temporal contexts to refine the prediction of SRFC. Specifically, TRFC firstly computes the temporal relations among the same regions across \textit{all input frames}.  Then, the temporal contexts are aggregated based on the computed temporal relations. Finally, TRFC uses the temporal contexts to help the feature recovery of occluded regions.

Different from the computationally expensive image completion network, the proposed RFC block is lightweight and imposes only a slight increase in model complexity. It can be readily inserted into any network. In our work, we integrate RFC block with ResNet-50~\cite{residual} to construct \textit{Region Feature Completion Network} (RFCnet).  \textbf{To facilitate the research on the video-based occluded reID, we reorganize DukeMTMC-VideoReID~\cite{dukereid} to form a large-scale dataset named Occluded-DukeMTMC-VideoReID}.  We demonstrate the effectiveness of RFCnet on both occluded and holistic reID benchmarks. Notably, as images can be regarded as single frame videos, the proposed SRFC can also be used for image person reID. Extensive experimental results show that our method performs favorably against state-of-the-arts. Especially on the occluded datasets, our method significantly outperforms state-of-the-art by about $10\%$ mAP.

In summary, the main contributions of our work lie in four aspects: (1) proposing to use feature completion to address the occluded person reID problem; (2) designing a SRFC and TRFC module that respectively captures spatial and temporal contexts to recover the features of occluded regions; (3) constructing a large-scale video occluded reID dataset Occluded-DukeMTMC-VideoReID to facilitate the research on occluded reID; (4) achieving superior performance on both occluded and holistic reID compared with state-of-the-art methods.

\section{Related Work}
\begin{figure*}[t]
\centering
   \includegraphics[width=0.9\linewidth]{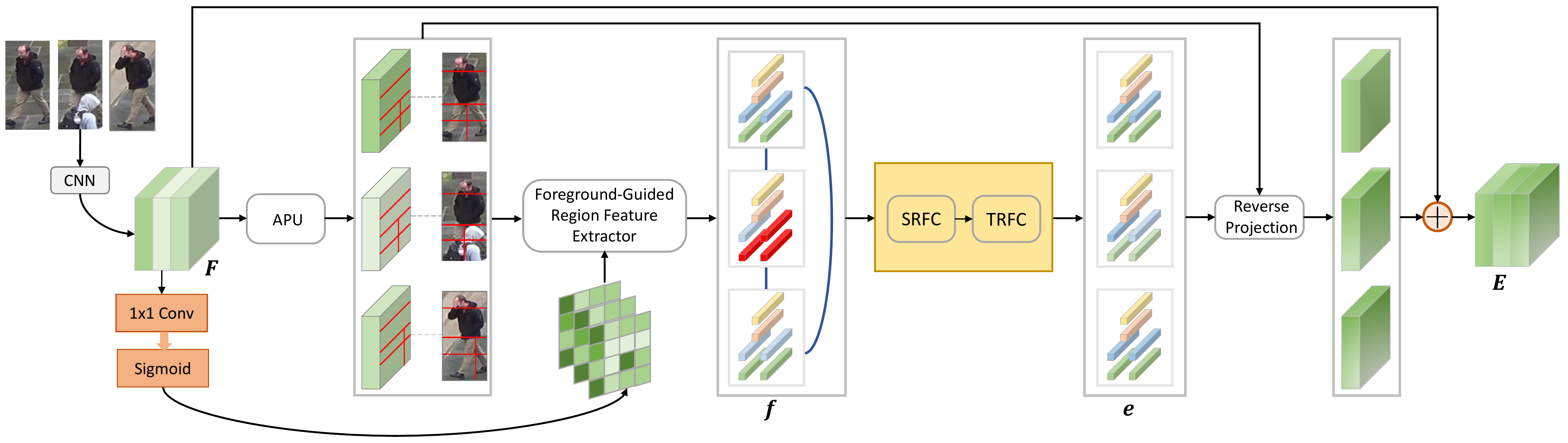}
   \caption{The overall architecture of \textit{Region Feature Completion} (RFC) block. The \textit{Adaptive Partition Unit} (APU) is firstly used to divide input feature maps into multiple regions. The \textit{Foreground-Guided Region Feature Extractor} is then employed to produce a body-aware probability map for each frame, which is then added to the divided regions to generated discriminative region features. Then, we sequentially feed the video region features into  \textit{Spatial Region Feature Completion} (SRFC) and \textit{Temporal Region Feature Completion} (TRFC) modules to recover the features of occluded regions. Finally, the \textit{Reverse Projection} operation projects the completed region features to original space, which makes RFC compatible with existing CNN architecture.}
\label{fig2}
\vspace*{-1em}
\end{figure*}
\subsection{Image Person ReID}
In recent years, deep learning methods\cite{background,smoothed,Learning,hard-aware,gu2019temporal,zhu2017fast,li2019unsupervised,wang2016person,spindle-net,short-term,point-to-set,multi-channel,KPM,MSL} show significant superiority on image person reID. A line of the work uses the siamese network which takes image pairs or triplets as the inputs. Li \textsl{et al.}~\cite{Cuhk} propose to input a pair of pedestrian images to a CNN and train the model with verification loss. Ding \textsl{et al.}~\cite{triplet-loss} further employ a triplet loss. Another line adopts identity classification models. Zheng \textsl{et al.}~\cite{mars} propose an identity discriminative embedding (IDE) that learns the features from multi-class person identification classification tasks. Further, Zhang \textsl{et al.}~\cite{zhang2019densely} train the model with a joint triplet and classification loss and achieve the state-of-the-art performance. 

Also, some works attempt to alleviate the occlusion for image person reID. Generally, previous methods typically either leverage external clues, or adopt part-to-part matching. \textit{External-clues guided methods} usually use the person poses or masks to remove the appearance of obstacles. For example, some works~\cite{background,mask-guided} use person masks to remove the background clutters at pixel-level. Further, Kalayeh \textsl{et al.}~\cite{semantic} integrate human semantic parsing to extract the features of body parts. Other works~\cite{pose-invariant,pose-driven,guo2019beyond} locate each body part using human landmarks and utilize the skeleton as an external cue to effectively relieve the occlusion interference. \textit{Part-based matching methods} employ a part-to-part matching strategy and aim at the cases where the target person is partially out of the camera's view, which is known as Partial person reID problem. The work~\cite{zheng2015partial} firstly defines the partial person reID, \textsl{i.e.}, a probe image contains only partial body of a person and the task is to match this partial observation with a gallery consisting of full-body images. And this work proposes to decompose the probe and gallery images into small local patches and performs patch-level local-to-local matching. Furthermore, Sun~\textit{et al.}~\cite{sun2019perceive} employ self-supervision for learning the probability of region visibility and then only match the shared body regions for a pair of images. Although above approaches can alleviate the occlusion issue, they completely discard the occluded parts and only rely on the non-occluded parts, which reduces the reliability of the retrieval results. In contrast, our RFC block can recover the features of occluded regions to obtain more discriminative representation.

\subsection{Video Person ReID}

Video person reID is an extension of the image setting, where video sequences are used instead of individual images. The powerful feature learning ability of CNN also inspires its application in video reID~\cite{TDL,K-liu,gu2019temporal,wang2016person,stepwise,ilids}. The key focus of existing studies for video person reID lies in the exploitation of temporal clues. The early works~\cite{RCN,jointly,See,snippet} use the optical flows to encode the short-term motion features between adjacent frames. Mclaughlin~\textsl{et al.}~\cite{RCN} employ a recurrent architecture to aggregate the frame-level representations and generate a sequence-level feature representation. Liao~\textsl{et al.}~\cite{V3DP}  use 3D convolution for spatial-temporal feature learning. Recently, some works~\cite{V3DP,VRSTC,GLTL} apply non-local blocks~\cite{non-local} to model long-term temporal dependencies. 

To handle the occlusion problem for video person reID, the approaches based temporal attention are gaining popularity. Liu \textsl{et al.}~\cite{QAN} predict a quality score for each frame to weaken the interference of occluded frames. Zhou \textsl{et al.}~\cite{See} propose a RNN temporal attention mechanism to select the most discriminative frames from video. Further, the works~\cite{jointly,diversity} employ spatial and temporal attention layers, where the spatial attention layer localizes the discriminative body parts for each frame and the temporal attention layer selects informative frames in the video. The attention based methods usually discard the occluded frames directly, causing the loss of valuable spatial and temporal information of videos. Our previous work~\cite{VRSTC} proposes an image completion network for recovering the occluded pixels. In this work, our RFC block performs a feature completion of human body regions, which can be seamlessly integrated to reID network for end-to-end training. 

\subsection{Occluded Person ReID} Although some previous works attempt to solve the occlusion issue, the occluded reID task lacks a precise definition and benchmarks. Recently, the work~\cite{miao2019pose} introduces an image occluded reID dataset and clearly defines the occluded reID problem, in which \textit{all probe images are occluded and gallery images contain both holistic (non-occluded) and occluded images}. The task is more challenging since at least one occluded image exists when matching. Zhou~\textit{et al.}~\cite{zhuo2018occluded} use a binary classification loss to distinguish the occluded images from holistic ones and an identification loss to force the model to focus on non-occluded body parts. He~\textit{et al.}~\cite{he2019foreground} propose to reconstruct the feature map of the occluded pedestrian from the holistic ones. They also utilize the person masks to assign the occlusion parts with smaller weights to overcome the occlusion problem. Miao~\textit{et al.}~\cite{miao2019pose} propose a pose guided feature alignment mechanism to match the shared body parts of probe and gallery images based on the pose landmarks. Specifically, they use a pre-defined threshold of landmark confidence to determine whether a body part is occluded or not and filter out the information of occluded parts in the matching stage. 

However, most existing occluded person reID methods only focus on the pedestrian images. In this work, we propose a framework for video occluded person reID. Since images can be regarded as singe frame videos, our method can also be used for image reID. Also, unlike previous works which only discard the occluded parts, our RFC block can use the spatial-temporal contexts to automatically recover the features of occluded parts for more accurate person retrieval.

\section{Region Feature Completion Block}

The architecture of our RFC block is illustrated in Fig.~\ref{fig2}. Suppose a convolutional video feature map $F\in\mathbb{R}^{T\times H\times W \times D}$ is given, where $T$, $H$, $W$ and $D$ denote the frame number, the height, the width and the channel number of the feature map respectively. We first use an \textit{Adaptive Partition Unit} to adaptively divide input feature maps into different regions. A \textit{Foreground-Guided Region Feature Extractor} is then employed to produce a body-aware probability map for each frame, which is then added to the divided regions to generate discriminative region features. Then, we feed the video region features into two completion modules, \textit{Spatial RFC} and \textit{Temporal RFC}. The two completion modules employ complementary clues, focusing on spatial and temporal respectively. Considering this, two modules can be placed in parallel or sequential manner. We find that the sequential arrangement gives a better result than a parallel arrangement. We argue that the sequential arrangement enables a progressive completion where one module gives a coarse prediction and the other refines the initial prediction. 
Finally, we project the completed region features to the original space to make above building blocks compatible with existing CNN architecture.

\subsection{Adaptive Partition Unit} 

In order to remove the corruption of occlusion, existing methods~\cite{RQAN,he2019foreground} usually conduct \textit{fixed partition} on the convolutional feature maps and then analyze whether each divided region contains occlusions or not. However, as pointed by the works~\cite{PCB,pose-invariant,PNGAN}, the fixed partition is prone to the spatial misalignment of pedestrians.  Sun~\textsl{et al.}~\cite{PCB} propose refined part pooling (RPP) to improve the fixed partition. But RPP relies on the appearance information that may fail to accurately locate the body parts under occlusion. In particular, when the target person is occluded by another
pedestrian, the part region generated by RPP may contain the corresponding part of the other person,
resulting in an inaccurate part feature.

\begin{figure}[t]
\centering
   \includegraphics[width=1.0\linewidth]{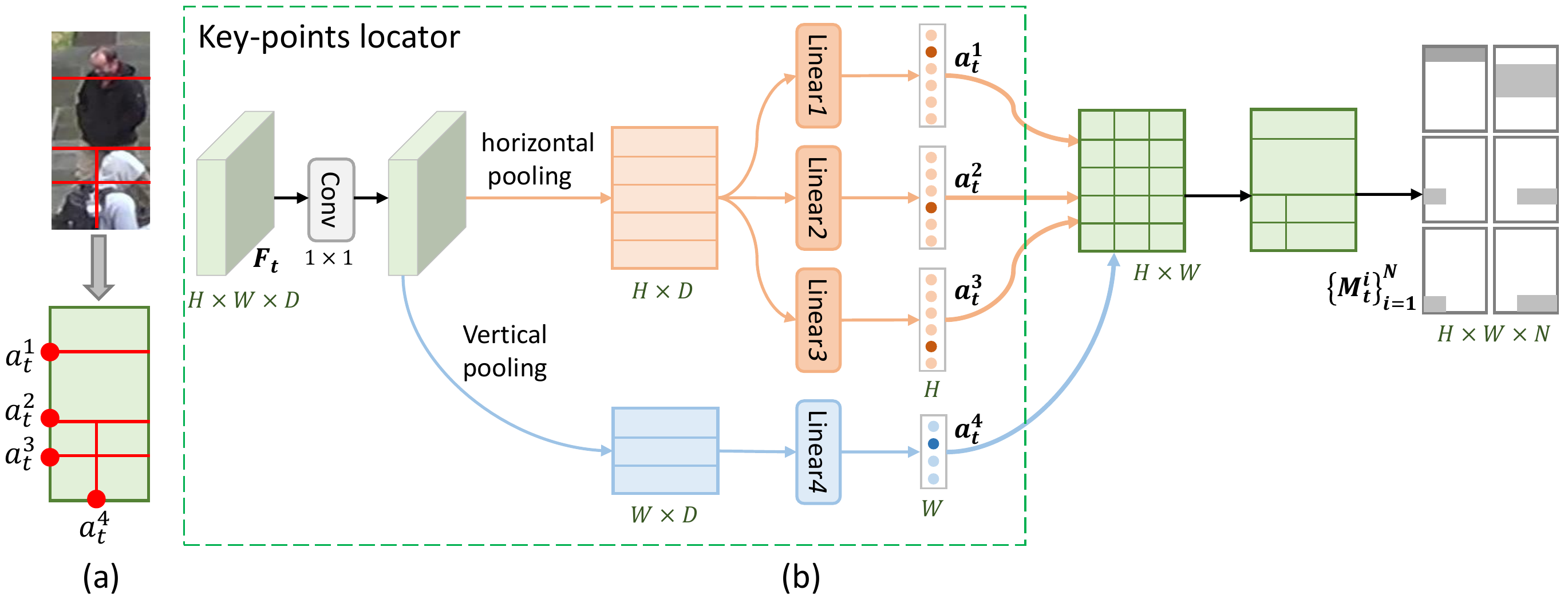}
   \caption{\textit{Adaptive Partition Unit} (APU) (a) An example of APU. (b) The architecture of APU. In $M_t$, the values of gray-color pixels are $1$ and the values of white-color pixels are $0$.}
\label{APU}
\end{figure}

\begin{figure*}[t]
\centering
   \includegraphics[width=0.9\linewidth]{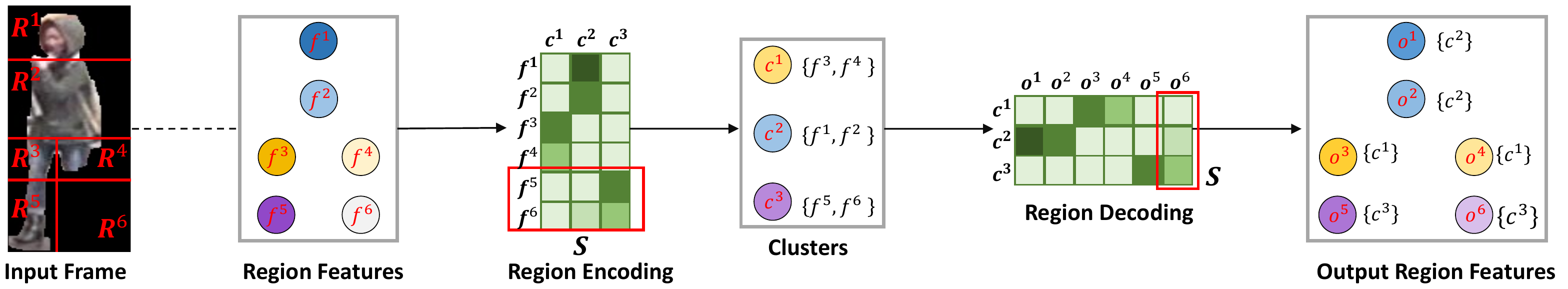}
   \caption{An example of region encoding and decoding of SRFC module. SRFC processes each frame independently. For simplicity, we omit the subscript $t$ and denote region feature $f_t$ as $f$. In the region-encoding, the features of divided regions $\{f^i\}_{i=1}^N$ are aggregated to a few clusters $\{c^k\}_{k=1}^K$ by an assignment matrix $S\in\mathbb{R}^{N\times K}$. $S_{ik}$ indicates the probability of assigning  region $R^i$ to  $c^k$, and \textit{a darker color indicates a higher probability value in this figure}. SRFC constrains to assign the regions with similar appearance or close position to a cluster. So each cluster gathers the features of most correlated regions and usually represents a body part. In this figure, $c^1$, $c^2$ and $c^3$ correspond to the pants, upper-body  and boots respectively. In the region-decoding, $\{c^i\}_{i=1}^K$ are distributed to output region features $\{o^i\}_{i=1}^N$ by $S$. So the occluded region can use the correlated cluster to recover its feature. For example, $R^6$ can use $c^3$ to recover its feature to represent the boots.}
\label{SFC-motivation}
\end{figure*}

To this end, we design an \textit{adaptive partition unit} which can adaptively divide input feature maps into different regions corresponding to specific body parts. As occlusion typically occurs on the lower part of the pedestrians, we conduct a fine-partition on the lower body. Specifically, we define $N$ regions on the person images, which are associated to different body parts, namely head, upper body, upper-left leg, lower-left leg, upper-right leg, lower-right leg. Then on each frame feature map $F_t$, APU appends a key-points locator to discover the pre-defined regions on $F_t$, and then generates a respective  mask for each region.

\textbf{Key-points Locator}. As shown in Fig.~\ref{APU} (a), four key points, $\{(a_t^1, 0)$, $(a_t^2, 0)$, $(a_t^3, 0)$, $(0, a_t^4)$\}, are enough to obtain aforementioned partition for a person image. To this end, we design a key-points locator to predict the locations of the four key points on $F_t$, which is considered as a \textit{multi-class classification} problem. For $a_t^1$ as an example, the locator predicts a probability distribution $p_t^1\in\mathbb{R}^{H}$, where $(p_t^1)_h$ is the probability of $a_t^1=h$. Then $a_t^1$ can be obtained by taking the index of the maximum value of $p_t^i$.

Specifically, as shown in Fig.~\ref{APU} (b), we first process $F_t$ with a $1\times 1$ convolution to reduce the channel dimension. Then for each column-wise key point $a_t^i \left(i\in\{1,2,3\}\right)$, we perform average pooling along the horizontal direction to obtain a column-wise descriptor. After that, we feed the flatted column-wise descriptor to a linear layer followed by a Softmax function to generate corresponding probability distribution $p_t^i\in\mathbb{R}^{H}$. Finally, $a_t^i$ is achieved by taking the index of the maximum value of $p_t^i$ as $a_t^i = \arg \max(p_t^i)$. Similarly, we perform average pooling along the vertical direction to obtain a row-wise descriptor. Then we employ a linear layer on the flatted row-wise descriptor to produce the probability distribution $p_t^4\in\mathbb{R}^{W}$ and obtain $a_t^4$ as $\arg \max (p_t^4)$.  Notably, we use the human key points obtained by a pose estimator model~\cite{liang2018look} to guide the generation of $\{p_t^i\}_{i=1}^4$. The detailed implementation is described in Section~\ref{object-function}.

According to the generated key points,
each pixel $(F_t)_{hw}$, where $(h, w)$ denotes the spatial location of the pixel on $F_t$, can be classified into the pre-defined regions $\{R_t^i\}_{i=1}^N$:
\begin{equation}
 (F_t)_{hw} \in \left\{
\begin{aligned}
R_t^1 & \ (h \leq a_t^1) \\
R_t^2 & \ (a_t^1 < h \leq a_t^2)\\
R_t^3 & \ (a_t^2 < h \leq a_t^3 , w \leq a_t^4)\\
R_t^4 & \ (a_t^2 < h \leq a_t^3, w > a_t^4)\\
R_t^5 & \ (h > a_t^3, w \leq a_t^4 )\\
R_t^6 & \ (h > a_t^3, w > a_t^4 )\\
\end{aligned}
\right.
\end{equation}
Finally, we can obtain the region mask $M_t^i\in\mathbb{R}^{H\times W}$ for each region $R_t^i$ by setting the values of pixels in $R_t^i$ to $1$, as shown in Fig.~\ref{APU} (b).

Notably, we can simply replace APU with other methods~\cite{part-aligned,harmoniou,attention-aware} that can locate the body parts. However, compared to the unsupervised methods~\cite{part-aligned,harmoniou}, APU can more accurately locate the body parts with the guidance of external human pose information. And compared to the method~\cite{attention-aware} that requires to predict all human key points, APU  only predicts four coarse keypoints which is more robust to the pose detection noise. Therefore, we choose to use APU in RFC block.


\subsection{Foreground-Guided Region Feature Extractor} 
The Foreground-Guided Region Feature Extractor is used to extract the features for the divided regions of APU. The extractor uses a foreground map to guarantee the region features less corrupted by occlusion. As shown in Fig.~\ref{fig2}, we employ a $1\times 1$ convolutional layer followed by a Sigmoid function to produce a foreground probability map $I_t\in \mathbb{R}^{H\times W}$ for frame feature map $F_t$. We use the body masks obtained by a human parsing model~\cite{liang2018look} to guide the generation of $I_t$. The detailed implementation is described in Section~\ref{object-function}. With the guidance of body masks, the  foreground maps can assign relatively large values to foreground pixels while relatively small values to background and occluded pixels. Thus we can leverage $I_t$ to generate a discriminative region features $f_t \in \mathbb{R}^{N\times D}$,
\begin{equation}
f_t^i = \left(f_t\right)_i = \frac{M_t^i \odot I_t}{||M_t^i \odot I_t||_1} * F_t.
\end{equation}
Here  $f_t^i$ is the feature vector corresponding to the $i^{th}$ region, $\odot$ is the element-wise multiplication  operation, $||.||_1$ denotes the $L_1$ norm of the matrix and $*$ denotes the weighted sum operation.

\subsection{SRFC Module}
The foreground-guided extractor masks the appearance information of the occluders. So it can alleviate the feature corruption by the visual appearance of occluder. However, occlusion still leads to information loss of the target person. To this end, we propose SRFC that uses spatial information to complement the features of occluded regions.  SRFC processes the region feature of each frame independently. For simplicity, we omit the subscript $t$ in this subsection and denote the region feature $f_t$ as $f$.

As shown in Fig.~\ref{SFC-motivation}, SRFC consists of a region-encoder and region-decoder. In the region-encoder, the features of all divided regions $\{f^i\}_{i=1}^N$ are aggregated to a few clusters $\{c^k\}_{k=1}^K$ ($K$$<$$N$). The region encoding is implemented by an assignment matrix $S\in\mathbb{R}^{N\times K}$. Here $S_{ik}$ indicates the probability of assigning region $R^i$ to cluster $c^k$, and $S_i\in\mathbb{R}^{K}$ is denoted as the assignment vector of $R^i$. The assignment vector is generated based on the \textbf{appearance} and \textbf{position} information of corresponding region. \textbf{The key is that if different regions have similar appearances or close positions, the region-encoder outputs similar assignment vectors for these regions}. So region-encoder can assign the correlated regions (with similar appearances or close positions) to a cluster. In this way, each cluster gathers the features of most correlated regions and usually represents a large body part. For example,  $c^1$, $c^2$ and $c^3$ correspond to features of pants, upper-body and boots respectively in Fig.~\ref{SFC-motivation}. 

Then  the region-decoder distributes the clusters $\{c^k\}_{k=1}^K$ to output region features $\{o^i\}_{i=1}^N$ based on $S$.  So for the occluded region, the region-decoder can use the correlated cluster to recover its feature. As shown in Fig.~\ref{SFC-motivation}, the region-encoder relies on position information to assign $R^6$ and adjacent $R^5$ to a cluster $c^3$, while region-decoder predicts the feature of $R^6$ mainly based on $c^3$. With the cluster as intermediary, the information from $R^5$ can be propagated to $R^6$ to recover its feature representation. 
\begin{figure*}[t]
\centering
   \includegraphics[width=0.9\linewidth]{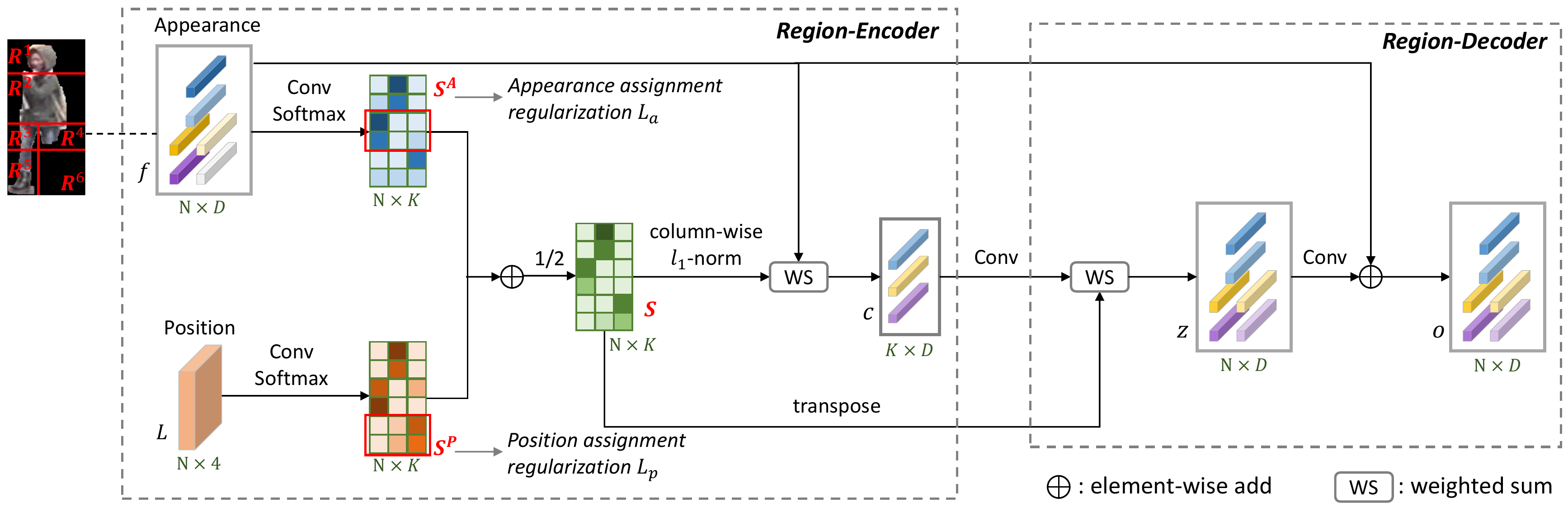}
   \caption{The architecture of SRFC module.  It is mainly composed of a region-encoder and a region-decoder. The \textbf{region-encoder} maps the input $N$ region $\{R^i\}_{i=1}^N$ to $K$ clusters $\{c^k\}_{k=1}^K$. Specifically, it firstly takes the region appearance feature $f$ and position encoding $L$ as inputs, and outputs an assignment matrix $S\in\mathbb{R}^{N\times K}$. Here $S_{ik}$ indicates the probability of assigning region $R^i$ to $c^k$. Then each cluster $c^k$ is obtained by  summing $\{f^i\}_{i=1}^N$ weighted by the normalized $k^{th}$ column of $S$. With the proposed \textit{appearance} and \textit{position assignment regularization}, the highly correlated regions (with similar appearances or close positions) tend to have consistent assignment vectors. In this way, each cluster can aggregate the features of most correlated regions to describe a body part. The \textbf{region-decoder} then uses the clusters to predict new region feature $z\in\mathbb{R}^{N\times D}$. Here $z^i$ is obtained by summing $\{c^k\}_{k=1}^K$ weighted by the $i^{th}$ row of $S$. In this way, the occluded regions can use the information of the most correlated cluster to recover its feature.  Finally, we adopt a residual learning strategy to ease the completion task.}
\label{SFC}
\vspace*{-1em}
\end{figure*}

\textbf{Region-Encoder.} \quad 
The detailed structure of SFCnet is illustrated in Fig.~\ref{SFC}.  SRFC consists of a region-encoder and a region-decoder. The region-encoder learns an assignment matrix to map the region features $f$ to a set of clusters $\{c^k\}_{k=1}^{K}$. 

\textbf{Appearance Information.} \quad  We first use the region feature $f\in\mathbb{R}^{N\times D}$ to generate an appearance assignment matrix $S^A\in\mathbb{R}^{N\times K}$. It is implemented by a convolutional layer followed by a softmax function as,
\begin{equation}
S^A = \text{softmax}\left(W^{a} * f\right),
\label{eq3}
\end{equation}
where $*$ is the convolutional operation, $W^a\in\mathbb{R}^{1\times1\times D\times K}$ is the convolutional filter weight.  The softmax function is applied in row-wise fashion to guarantee $\sum_{k=1}^{K} S^A_{ik}=1$. And $S^A_{ik}$  represents the probability of assigning region $R^i$ to cluster $c^k$.

\textbf{Constraining to assign similar-appearance regions to a cluster.} $S^A$ aims to assign the regions with similar appearance to a cluster. In other words, if the features of region $R^i$ and $R^j$ have a high appearance similarity, the corresponding appearance assignment vectors $S^A_i$ and $S^A_j$ should be highly similar.  To achieve this, we introduce an \textit{appearance assignment regularization} term, which constrains the similarities of appearance assignment vectors to be consistent with the similarities of corresponding region features,
\begin{equation}
L_{a} = \sum_{i}^N\sum_{j}^N ||\textit{sim}\left(S^A_i, S^A_j\right) - \textit{sim}\left(f^i, f^j\right) ||_{1}.
\label{eq4}
\end{equation}
Here  $||.||_1$ denotes $l_1$ distance, and $\textit{sim}$ measures the cosine similarity, which is defined as $\textit{sim}(x,y)$$=$$\frac{x^Ty}{||x||_2||y||_2}$. With $L_a$ constraint, the region-encoder tends to produce consistent appearance assignment vectors for appearance-similar regions.  Therefore, $S^A$ can assign the regions with similar appearances to a cluster.

\textbf{Feature completion of \textit{partially} occluded regions with $\mathbf{S^A}$.} $S^A$ is conducive to the feature completion of partially occluded regions. For example, as shown in Fig.~\ref{SFC}, the region $R^4$ is partially occluded where the remaining visible area has a similar appearance to $R^3$, With $L_a$ constraint, $S^A$ can assign $R^4$ and $R^3$ to a cluster. In this way, the information from $R^3$ can be propagated to $R^4$, thereby helping its feature completion. 

\textbf{Position  Information.} \quad 
To complete the fully occluded regions, we additionally utilize the position information to connect adjacent regions. Concretely, we firstly encode the position information of region $R^i$ with a vector $(y_i,x_i,h_i,w_i)$, where $(y_i, x_i)$ denotes the spatial coordinate of the region center, and $h_i$ and $w_i$ denote the height and width of the region respectively. Then, we define the position encoding of divided $N$ regions as $L\in\mathbb{R}^{N\times 4}$. Finally, the region-encoder produces a position assignment matrix $S^P \in \mathbb{R}^{N\times K}$ as,
\begin{equation}
S^P=\text{softmax}\left(W^p_{2}*\text{max}\left(0, W^p_1*L\right)\right),
\label{eq5}
\end{equation}
where $W^p_1\in\mathbb{R}^{1\times 1\times 4\times D}$ and $W^p_2\in\mathbb{R}^{1\times 1\times D\times K}$ correspond to two learnable convolutional kernels. Analogously to Eq.~\ref{eq3}, the softmax function is applied in row-wise fashion to guarantee $\sum_{k=1}^{K} S^P_{ik}=1$.

\textbf{Constraining to assign close-position regions to a cluster.} 
$S^P$ aims to assign the close regions  to a cluster. In other words, if the regions $R^i$ and $R^j$ are close, the corresponding position assignment vectors $S^P_i$ and $S^P_j$ should be highly similar.  To achieve this, we firstly measure the position similarity of two regions $R^i$ and $R^j$ as:
\begin{equation}
ps(R^i,R^j)=1-2\sqrt{\left(\frac{y_i-y_j}{H}\right)^2+\left(\frac{x_i-x_j}{W}\right)^2}.
\label{eq6}
\end{equation}
It can be inferred from Eq.~\ref{eq6} that $ps(R^i,R^i)=1$ and $ps(R^i,R^j)\in [-1,1]$. Thus the position similarity is comparable to the cosine similarity. To this end, we introduce a \textit{position assignment regularization} term, which constrains the cosine similarity of position assignment vectors to be consistent with the position similarity of corresponding regions,
\begin{equation}
L_{p} = \sum_{i}^N\sum_{j}^N\left|\textit{sim}\left(S^P_i, S^P_j\right) - \textit{ps}\left(R^i, R^j\right)\right|_{1}.
\label{eq7}
\end{equation}
With $L_p$ constraint, the region-encoder tends to produce consistent position assignment vectors for close regions. Therefore, $S^P$ can assign the close regions to a cluster. 

\textbf{Feature completion of \textit{fully} occluded regions with $\mathbf{S^P}$.} 
$S^P$ is conducive to complete the \textit{fully occluded regions}. For instance, as shown in Fig.~\ref{SFC}, the region $R^6$ is fully occluded with no appearance clues. In this case,  the region-encoder can use the position information. With $L_p$ constraint, $S^p$ can map $R^6$ and the most adjacent region $R^5$ to a cluster. Thus the information from $R^5$ can be propagated to $R^6$ to recover its feature representation.

\textbf{Encoding Matrix.} \quad
The assignment matrix $S$ integrates the appearance with position information as $S=(S^A+S^P)/2$. It can be inferred that $\sum_{k=1}^K S_{ik}=\left(\sum_{k=1}^K S^A_{ik} + \sum_{k=1}^K S^P_{ik}\right)/2 = 1$. So $S_{ik}$ can represent the probability of assigning region $R^i$ to $c^k$.

Based on $S$, we can generate the bag of clusters $\{c^k\}_{k=1}^K$, where $c^k$ is obtained by aggregating the region features weighted by the $k^{th}$ column of $S$. This inspires us to develop an attention-based feature encoding operation. We further apply a \textit{column-wise $l_1$-normalization} on $S$ to generate an encoding matrix $A\in\mathbb{R}^{N\times K}$ and perform region encoding as:
\begin{equation}
c^k = \sum_{i=1}^{N} A_{ik}f^i, \ \text{where}, \ A_{ik}=\frac{S_{ik}}{\sum_{j=1}^N S_{jk}}
\label{eq8}
\end{equation}  
In this way, each cluster gathers the most correlated regions. With the cluster as an intermediary, we can build the connection between occluded regions and long-distance non-occluded regions, thereby helping the feature completion of occluded regions.

It can be inferred from Eq.~\ref{eq8} that the resulting cluster $c^k$ linearly aggregates the features from different regions. But the linear aggregation may be insufficient to provide the clusters powerful representation ability. Therefore, we add a channel transformer layer with a $1\times1\times D\times D$ convolutional kernel on $c^k$ to enhance its representation.

\textbf{Region-Decoder.} \quad
The region-decoder performs an inverse operation of the region-encoder. It distributes the clusters to each region based on assignment matrix $S$. Specifically, we first transform $S$ to obtain the decoding matrix $B\in\mathbb{R}^{K\times N}$. Then, the clusters are transformed to $z\in\mathbb{R}^{N\times D}$ as,
\begin{equation}
z^i = \sum_{k=1}^K B_{ki}c^k, \ \text{where}, \ B_{ki}=S_{ik}
\label{eq10}
\end{equation}
The reuse of $S$ can ensure the consistency of the encoding and decoding process. That is, a higher $S_{ik}$ tends to produce a higher $A_{ik}$ (Eq.~\ref{eq8}) and  $B_{ki}$ (Eq.~\ref{eq10}).  So if the region-encoder assigns a occluded region $R^i$ to cluster $c^k$, the region-decoder will predict the feature of this region from $c^k$. Since $c^k$ aggregates the features of correlated non-occluded regions, the occluded region can sense the correlated part information to recover its feature.

Finally, a channel transformer layer is attached on region-decoder to update $z$. Following image completion~\cite{context}, we adopt a residual learning strategy to ease the completion task, which is defined as $o = f + z$. In a similar way, we can obtain the completed region feature of each frame, forming the output $\{o_1, \dots, o_T\}$ for the input sequence.

\begin{figure}[t]
\centering
   \includegraphics[width=1.0\linewidth]{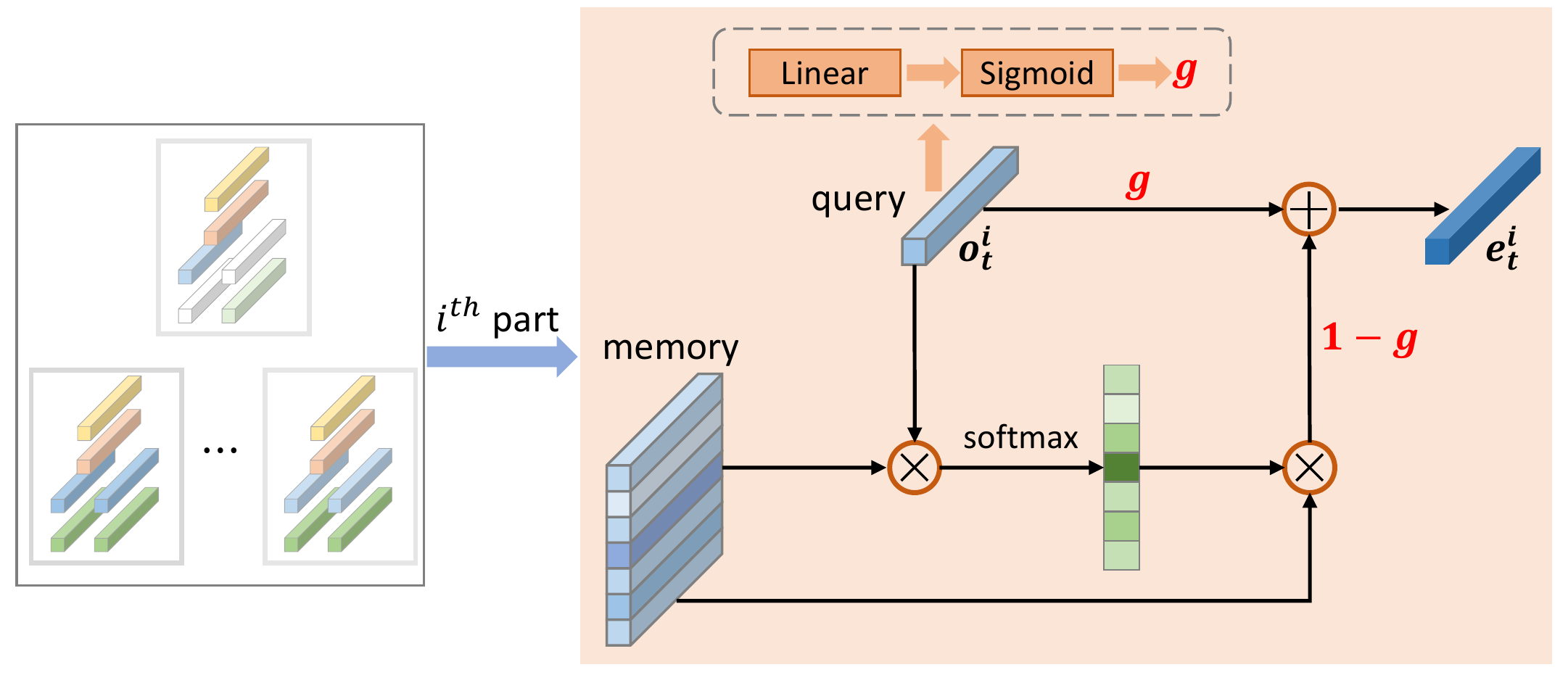}
   \caption{The architecture of (TRFC) module for the $i^{th}$ region in $t^{th}$ frame. Operations for other regions are the same.}
\label{TFC}
\vspace*{-1em}
\end{figure}

\subsection{TRFC Module}
TRFC module uses long-term temporal clues to refine the prediction of SRFC. It is based on a  \textit{query-memory} attention mechanism, where the region being processed is the query and the corresponding regions of remaining frames are the memory. The intuition is that the attention will add temporal contexts from other frames to the query vector, to aid with the feature completion.  

As shown in Fig.~\ref{TFC}, TRFC respectively considers each region feature generated by SRFC ($o_t^i$) as the query. Corresponding, a probability map can be obtained regarding how well the query vector matches each item of the memory through dot-product similarity~\cite{self-attention}. The temporal context vector $v_t^i$ is obtained as the sum of the items in the memory weighted by their probabilities. This operation can be succinctly represented as,

\begin{equation}
\begin{aligned}
&\alpha_{k}^i = \frac{\exp{\left((o_t^i)^To_k^i\right)}}{\sum_{l\neq t} \exp{\left((o_t^i)^To_l^i\right)}}, \\
&v_t^i = \sum_{k\neq t}\alpha_k^io_k^i.
\end{aligned}
\end{equation}
As last, we use a gate schema to control the required balance of how much temporal contexts should be passed to a query. Specifically, the query is updated by summing with the context vectors through a gating weight $g$,
\begin{equation}
\begin{aligned}
&g_t^i = \text{Sigmoid}(Wo_t^i+b), \\
&e_t^i = g_t^i \odot o_t^i + (1 - g_t^i) \odot v_t^i,
\end{aligned}
\end{equation}
where $(W,b)$ are trainable weights and bias variable, $\odot$ is element-wise multiplication. In this way, the query vector can be adaptively aggregated with the temporal contexts. Notably, we constrain $W$ to be positive by a ReLU function. Since RFC block uses the foreground-guided extractor to mask the occluded feature units, the features of occluded regions tend to have small values. With the linear transformation of positive weight $W$, TRFC tends to output a small gating weight $g$ for occluded region. Correspondingly, TRFC can assign more weight ($1-g$) to the temporal context vector for occluded region. In this way, the feature of occluded region can be better recovered to represent the target person. 

In a similar way, we can obtain the completed feature for each region, forming the output $e_t = [e_t^1, \dots, e_t^N]^T$ for the $t^{th}$ frame. 

\subsection{Embedding with Networks} 

To make above building blocks compatible with existing CNN architecture, we reuse the region mask $M_t$ to project the completed region feature $e_t$ to original space. Specifically, we first reshape $M_t$ to $\mathbb{R}^{N \times HW}$. Then we perform matrix multiplication between the transpose of $M_t$ and $e_t$ and reshape the results to obtain the updated feature $F'_t\in\mathbb{R}^{H \times W \times D}$. As last, we adopt a residual learning to ease the training process, which is defined as:
\begin{equation}
E_t = BN(F'_t) + F_t.
\label{eq9}
\end{equation}
 Here, BN is a batch normalization~\cite{BN} layer to adjust the scale of $F'_t$ to the initial $F_t$.

\section{Network Design for RFC}
\subsection{Network Architecture}
The proposed RFC blocks can be easily plugged into existing CNNs to form the RFCnet. In this work, we use ResNet50~\cite{residual} pre-trained on ImageNet~\cite{imagenet} as the backbone network.  We remove the last down-sampling operation, following~\cite{PCB} to enrich the spatial granularity. So the final extracted features can better describe the target person. We denote the architecture as \textit{modified ResNet50}.  For video reID, a temporal aggregation layer, which outputs the mean of all frame features of the input sequence, is added on the backbone to obtain a single feature for the sequence. 

RFC can be inserted at any depth of networks. Considering the computational complexity, we only place it at the bottlenecks of models where the downsampling of feature maps occurs. On one hand, different from previous works~\cite{he2019foreground,miao2019pose} that only consider occlusions in the end, RFC blocks are added in the earlier stages, which can alleviate occlusion corruptions at the bottom layers. On the other hand, multiple RFC blocks can be located at different stages. They can hierarchically complete the features of occluded parts, which further enhances the ability to deal with occlusion. Finally, we insert RFC blocks to stage$_2$ and stage$_3$ of the backbone network to form our RFCnet.  Notably, the image data only contains one frame, which can not use TRFC module. So we remove the TRFC module when applying RFC block for image reID.

\begin{table*}[t]
\caption{The tracklets ratio with a certain fraction of occluded frames on query and gallery set of Occluded-DukeMTMC-VideoReID.}
\centering
\small
\begin{tabular}{ l | c | c| c | c| c | c | c| c| c} 
\hline
Fraction of occluded frames ($\%$) & 0& 0-$30$ & $30$-$40$ & $40$-$50$& $50$-$60$ & $60$-$70$ & $70$-$80$ & $80$-$90$ & $90$-$100$ \\
\hline
Tracklets ratio on query &$0\%$ &$0\%$ &$1.1\%$ &$2.1\%$ &$4.0\%$ & $8.4\%$ &$13.8\%$ & $33.2\%$ & $33.4\%$\\
Tracklets ratio on gallery &$31.0\%$ &$24.3\%$ &$1.3\%$ &$1.4\%$ &$2.0\%$ & $3.6\%$ &$5.9\%$ & $12.4\%$ & $18.1\%$\\
\hline
\end{tabular}
\label{tab-data}
\end{table*}

\begin{table*}[t]
\caption{The ratio of video frames with a certain fraction of occlusion on query and gallery set of Occluded-DukeMTMC-VideoReID.}
\centering
\small
\begin{tabular}{ l | c | c| c | c| c | c| c| c} 
\hline
Occlusion fraction ($\%$) &$0$-$10$ & $10$-$20$ & $20$-$30$& $30$-$40$ & $40$-$50$ & $50$-$60$ & $60$-$70$ & $70$-$100$ \\
\hline
Frames ratio on query & $13.5\%$ &$11.1\%$ &$12.4\%$ &$14.8\%$ & $14.9\%$ &$15.6\%$ & $11.1\%$ & $6.6\%$\\
Frames ratio on gallery &$13.3\%$ &$11.2\%$ &$12.3\%$ &$14.7\%$ & $14.4\%$ &$14.6\%$ & $11.2\%$ & $8.3\%$\\
\hline
\end{tabular}
\label{tab-data1}
\end{table*}

\begin{table}[!t]
\small
\centering
\caption{Occluded dataset details. We respectively use the abbreviation Occluded-Duke and Occluded-Duke-Video to represent the Occluded-DukeMTMC and Occluded-DukeMTMC-VideoReID datasets for convenience.}
\vspace*{-0.5em}
\label{duke}
\begin{center}
\begin{tabular}{c | c c}
\hline
Dataset & Occluded-Duke &  Occluded-Duke-Video  \\
\hline
Train (ID/Images) & 702/15,168  &702/292,343\\
Gallery  (ID/Images) &1,110/17,661 &1,110/281,114\\
Query (ID/Images) &519/2,210 &661/39,526\\
\hline
\end{tabular}
\end{center}
\label{occ-data}
\vspace*{-1.0em}
\end{table}

\subsection{Objective Functions.} 
\label{object-function}

In this part, we describe the objective function to train the network. The total loss of our network is denoted as:
\begin{equation}
L_{all} = (L_{ce} + L_{tri}) + \lambda_{1}L_{k}  + \lambda_{2}L_{f} + \lambda_{3} (L_{a} + L_{p}).
\label{eq17}
\end{equation}
Here $L_{ce}$ is the Cross Entropy Loss, $L_{tri}$ is the Batch Triplet loss, $L_k$ denotes the Key-points Loss, $L_f$ denotes the Foreground Maps Loss, and  $L_{a}$ (Eq.~\ref{eq4}) is the appearance assignment regularization term  and $L_{p}$  (Eq.~\ref{eq7}) is the position assignment regularization term. $\lambda_1$, $\lambda_{2}$ and $\lambda_3$  are the hyperparameters to balance the effects of different loss functions.

\textit{Cross Entropy Loss ($L_{ce}$).} The supervised loss is used to calculate the classifier error among the identities. The number of nodes in the softmax layer depends on the number of identities of the training set. 

\textit{Batch Triplet Loss ($L_{tri}$).}
To reduce intra-class variation and increase inter-class variation, we additionally adopt a batch hard triple loss~\cite{Triplet}. For each sample in a batch, it only selects the hardest positive and hardest negative samples within the batch to from the triples. We refer the readers to ~\cite{Triplet} for a more detailed description of the batch triplet loss.

\textit{Key-points Loss ($L_k$).} 
The key-points locator in APU aims to predict the location of the key-points. 
The label of predicted key-point probability $\{p_t^i\}_{i=1}^4$ is a one-hot vector $\{l_t^i\}_{i=1}^4$ obtained by a pose estimator model\footnote{We use the vertical position of detected key-points \textit{shoulder}, \textit{hip} and \textit{knee} as the label of $p^1$, $p^2$ and $p^3$, respectively. Also, we use the horizontal position of detected key-points \textit{knee} as the label of  $p^4$. }~\cite{liang2018look}. The key-points loss is  defined as,
\begin{equation}
L_k = \frac{1}{4T}\sum_{t=1}^T\sum_{i=1}^4  \textit{CE} \left(p_t^i, l_t^i\right),
\end{equation}  
where \textit{CE} denotes the cross entropy loss.

\textit{Foreground Maps Loss ($L_f$).} 
The foreground probability map (shown in Fig.~\ref{fig2}) aims to classify the background/occlusion part and the body part. We treat this problem as a binary classification problem. The label of predicted foreground map $I_t\in\mathbb{R}^{H\times W}$ is determined by the person mask $G_t\in\mathbb{R}^{H\times W}$ obtained by a human paring model~\cite{liang2018look}. Then the foreground maps loss is given by a binary cross entropy loss,
\begin{equation}
\begin{aligned}
L_{f} = -\frac{1}{THW}\sum_{t=1}^{T} \sum_{i=1}^{H} \sum_{j=1}^{W} [(G_t)_{ij} \log((I_t)_{ij}) &\\
   + (1-(G_t)_{ij}) \log(1-(I_t)_{ij}) ].
  \end{aligned}
\end{equation}

\section{Occluded-DukeMTMC-VideoReID Dataset}
\label{Sec6}
To facilitate the research on the Occluded Person ReID problem, we reorganize the DukeMTMC-VideoReID dataset~\cite{dukereid} to form a large-scale \textbf{occluded video-based} person re-identification dataset, Occluded-DukeMTMC-VideoReID.

\subsection{Data Reorganization}
\label{sec6.1}
Following the collection process of Occluded-DukeMTMC~\cite{miao2019pose}, we re-split the DukeMTMC-VideoReID testing set so that it contains \textbf{$\textbf{100\%}$ occluded query tracklets}. Specifically, all query tracklets contain occlusion by manually selecting from both the gallery and query set of the original dataset. Therefore, there always exists at least one occluded tracklet when calculating the pairwise distance between query and gallery tracklets. 

There are three steps to generate query tracklets for Occluded-DukeMTMC-VideoReID: 1) annotate the frame which contains more than one person as an occluded frame. Also annotate the frame in which a person is occluded by obstacles (\textit{e.g} trees or cars) as an occluded frame. 2) select the tracklets containing occluded frames from the gallery and query set of the original dataset. 3) from each selected tracklet, we randomly select a sub-tracklet, of which above $1/3$ frames contain occlusion, as a query tracklet.

When constructing the training set, we manually remove $494$ tracklets from original  Occluded-DukeMTMC training set, because these $494$ tracklets contain exactly the same obstacles as in the testing set. Those tracklets may lead the model to ``remember'' these specific occlusion patterns in the testing set, which overestimates the generality of the trained model. 

In the end, the training set contains 1,702 tracklets covering 702 identities in total, the query set contains 661 tracklets of 661 identities  and the gallery set contains 2,636 tracklets of 1,110 identities. The dataset can be downloaded in \url{http://vipl.ict.ac.cn/database.php}.

\subsection{Properties of Occluded-DukeMTMC-VideoReID}
There are a few properties to make Occluded-DukeMTMC-VideoReID appealing. First, as shown in Tab.~\ref{occ-data}, it is the largest occluded person reID dataset to date. Second, previous datasets~\cite{miao2019pose,zhuo2018occluded} only focus on the image setting. Occluded-DukeMTMC-VideoReID is the first occluded dataset for video-based person reID. Third, there are rich variations in Occluded-DukeMTMC-VideoReID, including different viewpoints and large variety of obstacles, including cars, trees, bicycles and other persons.

Because we care about the models' performance on test set, we further give more statistical-orient analysis of the test set 
of Occluded-DukeMTMC-VideoReID dataset. \textit{Firstly}, we give the fraction of occluded tracklets on query$/$gallery set. As described in Section~\ref{sec6.1}, all query tracklets contain occlusion. So $100\%$ tracklets of the query set has occlusion. And we count that above $70\%$ tracklets of the gallery set has occlusion. \textit{Secondly}, we annotate the fraction of occluded frames in each tracklet of query$/$gallery set.\footnote{all annotation files can be download in \url{http://vipl.ict.ac.cn/database.php}.} Furthermore, we count the tracklets ratio with a certain fraction of occluded frames. As shown in Tab.~\ref{tab-data},  above $60\%$ tracklets have more than $80\%$ occluded frames on query set, and above $30\%$ tracklets have more than $80\%$ occluded frames on gallery set. It indicates that the proposed dataset contains serious occlusion, which can  well evaluate the performance of the models on occluded scenes. \textit{Thirdly}, we annotate the occlusion fraction of each video frame on query$/$gallery set. In particular, we annotate each frame with label $\{0,1,2,\dots,10\}$\footnote{``0'' denotes the frame without any occlusion, ``1'' denotes $0$-$10\%$ fraction of the frame is occluded, and so on.}. Further,  in all occluded frames, we count the frame ratio with a certain of occlusion on query and gallery set in Tab.~\ref{tab-data1}. We observe that about $65\%$ frames contain less than $50\%$ occluded regions, which shows the \textit{partial occlusion} is more common on the dataset.  

\section{Experiments}
\subsection{Datasets and Settings}

\begin{table*}[t]
\caption{Comparison with state-of-the-arts on on image occluded reID dataset, Occluded-DukeMTMC. The ``CE'' and ``Triplet'' indicate whether the methods use the cross-entropy loss and triplet loss to train. The 
``Key-points'' and ``Foreground'' indicate whether the methods rely on extra supervision information from human pose model and human parsing model.}
\small
\centering
\begin{tabular}{ l | c | c c |  c c | c  c  c c}
\hline
\multirow{2}*{Methods} & \multirow{2}*{ Backbone}  & \multicolumn{2}{c|}{Loss Function} & \multicolumn{2}{c|}{External Clues} & \multicolumn{4}{c}{Occluded-DukeMTMC} \\
\cline{3-10}
& &CE & Triplet &key-points &Foreground &mAP  &top-1 &top-5 &top-10 \\
\hline
Part Aligned~\cite{part-aligned} & GoogLeNet & $\times$ & \checkmark &$\times$ & $\times$ &20.2 & 28.8 & 44.6 & 51.0 \\
\textbf{RFCnet}  &GoogLeNet &$\times$ & \checkmark & $\times$ &$\times$ &\textbf{36.0} &\textbf{44.6} &\textbf{61.2} &\textbf{67.2} \\
\hline
SFR~\cite{Ling-biometric} &ResNet50 & $\times$ &\checkmark &$\times$ &$\times$ &32.0 & 42.3 & 60.3 & 67.3\\
\textbf{RFCnet}  &ResNet50 & $\times$ & \checkmark & $\times$ & $\times$ & \textbf{43.4} &\textbf{53.2} & \textbf{68.5} &\textbf{73.4} \\
\hline
Part Bilinear~\cite{Suh-part-aligned} & GoogLeNet &$\times$ & \checkmark & \checkmark &$\times$ & - & 36.9 &- & -\\
\textbf{RFCnet}  &GoogLeNet &$\times$ &\checkmark &\checkmark &$\times$ &\textbf{37.1} &\textbf{46.3} &\textbf{61.9} & \textbf{68.6}\\
\hline
HACNN~\cite{harmoniou} &Inception &\checkmark &$\times$ &$\times$ &$\times$ & 26.0 & 34.4 & 51.9 & 59.4 \\
\textbf{RFCnet}  &Inception &\checkmark &$\times$ &$\times$ &$\times$ &\textbf{37.1} &\textbf{41.0} &\textbf{56.8} & \textbf{62.1}\\
\hline
Random Erasing~\cite{zhong2017random} & ResNet50 & \checkmark & $\times$ & $\times$ &$\times$ &30.0 &40.5 &59.6 &66.8  \\
DSR~\cite{he2018deep} &ResNet50 &\checkmark &$\times$ & $\times$ & $\times$ &30.4 & 40.8 & 58.2 & 65.2 \\
Adver Occluded~\cite{adversarially} & ResNet50  &\checkmark & $\times$ &$\times$ & $\times$ & 32.2 &44.5 &- & - \\
PCB~\cite{PCB} & ResNet50 & \checkmark &$\times$ & $\times$ &$\times$ & 33.7 & 42.6 & 57.1 &62.9 \\
 PCB+RPP~\cite{PCB} &ResNet50 & \checkmark & $\times$ &$\times$ &$\times$ &35.0 & 46.8 & 61.1 &67.3\\
\textbf{RFCnet} & ResNet50  &\checkmark & $\times$ & $\times$ &$\times$  & \textbf{44.8} &\textbf{52.4} & \textbf{68.3} &\textbf{73.4} \\
\hline
FD-GAN~\cite{Fd-gan} &ResNet50  & \checkmark & $\times$ & \checkmark &$\times$ &- &40.8 & -&- \\
PGFA~\cite{miao2019pose} & ResNet50  & \checkmark & $\times$ &\checkmark &$\times$ &37.3 & 51.4 & 68.6 & 74.9\\
\textbf{RFCnet} &ResNet50  & \checkmark & $\times$ & \checkmark & $\times$  &\textbf{46.1} & \textbf{54.6}&\textbf{70.3} &\textbf{75.9} \\
\hline
\textbf{RFCnet} & ResNet50  & \checkmark &\checkmark & \checkmark & \checkmark & \textbf{54.5} & \textbf{63.9}  & \textbf{77.6} & \textbf{82.1} \\
\hline
\end{tabular}
\label{Tab1}
\end{table*}
We evaluate our method on an image occluded reID datasets, Occluded-DukeMTMC~\cite{miao2019pose}, a video occluded reID dataset, Occluded-DukeMTMC-VideoReID, four image holistic datasets, Market-1501~\cite{Market1501}, DukeMTMC-reID~\cite{Duke}, CUHK03~\cite{Cuhk} and MSMT17~\cite{msmt17}, and two video holistic datasets, MARS~\cite{mars} and DukeMTMC-VideoReID~\cite{dukereid}.

\textbf{Occluded-DukeMTMC} is selected from DukeMTMC-reID by leaving occluded images as query and filtering out the images with same occlusion mode in the training set. It is the largest occluded image reID dataset.  The details can be seen in Tab.~\ref{occ-data}. There are a large variety of obstacles, making it is more challenging.

\textbf{Occluded-DukeMTMC-VideoReID} is our proposed video occluded reID datasets. See Section~\ref{Sec6} for more details.

\textbf{Market-1501} is a large-scale dataset that contains $1,501$ identities. The dataset is split into two fixed parts: $12,936$ images from $751$ identities for training and $19,732$ images from $751$ identities for testing. 

\textbf{DukeMTMC} is a subset of the multi-target, multi-camera pedestrian tracking dataset~\cite{multicamera}. There are $36,411$ images belonging to $1,404$ identities. It contains $16,522$ training images of $702$ identities, $2,228$ query images of the other $702$ identities and $17,661$ gallery images.

\textbf{CUHK03} dataset contains 13,164 images of 1,467 identities. Each identity is observed by 2 cameras. CUHK03 offers both hand-labeled and DPM-detected bounding boxes, and we use the latter in this paper following~\cite{PCB}. We adopt the new training/testing protocol proposed in \cite{re-rank}, and denote CUHK03 as CUHK03-NP in the following part.

\textbf{MSMT17} is the largest image person reID dataset. The training set contains 32,621 images of 1,041 identities, and the testing set contains 93,820 images of 3,060 identities. From the testing set, 11,659 images are randomly selected as query images, and the others are used as gallery images.

\textbf{MARS} is the largest video reID benchmark with $1,261$ identities and $17,503$ sequences captured by $6$ cameras. It consists of $631$ identities for training and the remaining identities for testing. The bounding boxes are produced by DPM~\cite{DPM} detector and GMMCP tracker~\cite{dehghan2015gmmcp}, such that it provides a more challenging environment similar to real-world applications. 

\textbf{DukeMTMC-VideoReID} is a subset of the tracking dataset DukeMTMC for video reID. The dataset consists of 702 identities for training, 702 identities for testing, and 408 identities as distractors. In total there are $2,196$ videos for training and $2,636$ videos for testing.

\begin{table*}[t]
\caption{Comparison with state-of-the-arts on video occluded reID dataset, Occluded-DukeMTMC-VideoReID. For fair comparison, we reproduce the other methods which use ResNet50 as the backbone network.} 
\small
\centering
\begin{tabular}{l | c|  c c | c c | c c c  c}
\hline
\multirow{2}*{Methods}  & \multirow{2}*{Backbone}  & \multicolumn{2}{c|}{Loss Function} & \multicolumn{2}{c|}{External Clues} & \multicolumn{4}{c}{Occluded-DukeMTMC-VideoReID} \\   
\cline{3-10}
& &CE & Triplet & key-points & Foreground &mAP  &top-1 &top-5 &top-10 \\
\hline
RCN~\cite{RCN} & ResNet50 &\checkmark &\checkmark &$\times$ &$\times$ &62.4 &60.9 &83.3 &88.1\\
TriNet~\cite{Triplet} &ResNet50 &\checkmark &\checkmark &$\times$ &$\times$ &64.1 &63.1 &82.6 &87.4\\
STAN~\cite{diversity} &ResNet50 &\checkmark &\checkmark&$\times$ &$\times$&69.4 &69.4 &88.1 &91.7\\
QAN~\cite{QAN} &ResNet50 &\checkmark &\checkmark &$\times$ &$\times$ &74.8 &75.1 &90.6 &93.4\\
RQEN~\cite{RQAN} &ResNet50 &\checkmark &\checkmark&$\times$ &$\times$ & 74.9 &73.5 &90.5 &94.1\\
VRSTC~\cite{VRSTC} &ResNet50 &\checkmark &\checkmark& $\times$ &$\times$ &76.7 &76.9 &90.3 &94.2\\
\textbf{RFCnet} &ResNet50&\checkmark &\checkmark &$\times$ &$\times$ & \textbf{90.1} & \textbf{90.5} &  \textbf{98.6} &  \textbf{98.9} \\
\hline
\textbf{RFCnet} &ResNet50 &\checkmark &\checkmark &\checkmark &\checkmark & \textbf{92.0} & \textbf{93.0}  & \textbf{98.6} & \textbf{99.1}\\
\hline
\end{tabular}
\label{Tab2}
\end{table*}

\textbf{Implementation Details for Image ReID.} For image reID, the input images are resized to $256\times 128$. We use random flipping and random erasing~\cite{zhong2017random} with a probability of $0.5$ for data augmentation. The random erasing implicitly adds occluded images to the training set, which is conducive to the learning of our RFC block. The initial learning rate is set to $0.00035$ with a decay factor $0.1$ at every $20$ epochs. Adam~\cite{adam} optimizer is used with a minibatch size of $64$ for $60$ epochs training. The balance rate $\lambda_1$, $\lambda_2$ and $\lambda_3$ of loss function are  set to $0.1$, $0.5$ and $0.05$ respectively. 

\textbf{Implementation Details for Video ReID.} For video reID, when training, we randomly sample 4 frames with a stride of 8 frames from the original full-length video to form an input video clip. RFCnet is trained for 150 epochs in total, with an initial learning rate of $0.0003$ and reduced it with decay rate $0.1$ every 40 epochs. The batch size is set to $32$. Other setting and hyperparameters are the same as those in the experiments of image reID. Notably, the reason of using 4-frame clips when training is due to GPU memory limitations. In particular, when a larger input-clip length is used, we have to reduce the batch size due to GPU memory limitations. However, the small batch size may not be conducive to the optimization of RFCnet, resulting in performance degeneration.

In the test phase, for each video tracklet, we first split it into several 64-frame video clips. Then we extract the feature representation for each video clip and the final video feature is the averaged representation of all clips\footnote{The parameters of RFC block is independent of the length of input sequences.}. So RFC can explore a long-term temporal contexts when testing. After feature extraction, the cosine distance between the query and gallery features are computed for retrieval.

\textbf{Evaluation Metric.}
We use standard metrics as in most person reID literatures, namely Cumulative Matching Characteristic (CMC) curves and mean Average Precision (mAP), to evaluate the quality of different person reID models. All the experiments are performed in single query setting.

\begin{table*}[t]
\caption{Comparison with state-of-the-arts on Market-1501, DukeMTMC, CUHK03 and MSMT17 datasets.  ``KP'' denotes key-points and   ``F'' denotes Foreground.}
\vspace*{-0.5em}
\small
\centering
\begin{tabular}{l |c | c c | c c| c  c|c c|c c| c c  }
\hline
\multirow{2}*{Methods} &\multirow{2}*{ Backbone}  & \multicolumn{2}{c|}{ Loss Function}& \multicolumn{2}{c|}{ External Clues}  & \multicolumn{2}{c|}{Market-1501}  &\multicolumn{2}{c|}{DukeMTMC} &\multicolumn{2}{c|}{ CUHK03-NP} &\multicolumn{2}{c}{ MSMT17}\\   
\cline{3-14}
& & CE &Triplet &KP & F &mAP  &top-1 &mAP &top-1  &mAP  &top-1 &mAP &top-1\\
\hline
PCB~\cite{PCB} & ResNet50 & \checkmark &  $\times$ &  $\times$ &  $\times$ & 77.4 & 92.3 &66.1 & 81.8&54.2 &61.3 &- &-\\
AO~\cite{adversarially} & ResNet50 &\checkmark & $\times$ & $\times$ & $\times$ & 78.3 &86.5 &62.1 & 79.1 &56.1&54.6&-&-\\
PCB+RPP~\cite{PCB} &ResNet50 &\checkmark &$\times$ &$\times$&$\times$ &81.6 &93.8 &69.2 &83.3&57.5 &63.7&-&-\\
CASN~\cite{zheng2019re} &ResNet50 &\checkmark &$\times$ & $\times$ & $\times$ &82.8 &94.4  &73.7 &\textbf{87.7}&64.4&71.5&-&-\\
IAnet~\cite{IANet} &ResNet50 & \checkmark & $\times$ & $\times$ &$\times$ &83.1 &94.4  &73.4 &87.1&-&-&46.8&75.5\\
\textbf{RFCnet} &ResNet50 &\checkmark & $\times$ & $\times$ & $\times$ & \textbf{84.8} & \textbf{94.5} &\textbf{76.5} & \textbf{87.7} &\textbf{69.7} &\textbf{73.3} &\textbf{51.5} &\textbf{76.4}\\
\hline
PGFA~\cite{miao2019pose} &ResNet50 &\checkmark &  $\times$ &\checkmark &$\times$ &76.8 &91.2 &65.5 &82.6 &- &- &- &-\\
\textbf{RFCnet} & ResNet50 & \checkmark &  $\times$ & \checkmark &  $\times$ & \textbf{85.7} &\textbf{94.5} & \textbf{76.6} & \textbf{87.7} &\textbf{69.9} &\textbf{73.5} &\textbf{52.7} &\textbf{77.7}\\
\hline
VPM~\cite{sun2019perceive} &  ResNet50 & \checkmark & \checkmark & $\times$  & $\times$ &80.8 &93.0 &72.6 &83.6 &- &- &- &-\\
BDB~\cite{dai2019batch} & ResNet50 &\checkmark & \checkmark & $\times$ & $\times$ &84.3 & 94.2 & 72.1 & 86.8 &69.3 &72.8 &- &-\\
Pymamid~\cite{zheng2019pyramidal} &ResNet50 & \checkmark & \checkmark & $\times$ & $\times$&\textbf{88.2} & \textbf{95.7} &79.0 & 89.0 &74.8&78.9&-&-\\
\textbf{RFCnet} & ResNet50 & \checkmark & \checkmark & $\times$ & $\times$ & \textbf{88.2} & 95.1 & \textbf{80.1} & \textbf{89.5} &\textbf{76.5} &\textbf{79.9} &\textbf{59.2} &\textbf{81.7}\\
\hline
FPR~\cite{he2019foreground} & ResNet50 & \checkmark & \checkmark & $\times$  & \checkmark &86.6 &\textbf{95.4} &78.4 &88.6 &72.3 &76.1&-&-\\
\textbf{RFCnet} &ResNet50 & \checkmark & \checkmark & $\times$ &\checkmark &  \textbf{88.6} & 95.1 &  \textbf{80.6} &\textbf{89.9} &\textbf{77.3} &\textbf{80.9}&\textbf{59.3} &\textbf{81.9}\\
\hline
DSA~\cite{zhang2019densely} &ResNet50 & \checkmark& \checkmark &\checkmark  &$\times$ & 87.6  &\textbf{95.7}  & 74.3 & 86.2 &73.1 &78.2 &- &-\\
\textbf{RFCnet} &ResNet50 &\checkmark &\checkmark &\checkmark &$\times$ & \textbf{88.7} &95.1 &\textbf{80.5} &  \textbf{90.0} &\textbf{76.5} &\textbf{80.0} &\textbf{59.6} &\textbf{81.9} \\
\hline
\textbf{RFCnet} &ResNet50 &\checkmark &\checkmark &\checkmark &\checkmark&  \textbf{89.2} & 95.2  &\textbf{80.7} &\textbf{90.7} &\textbf{78.0} &\textbf{81.1}&\textbf{60.2}&\textbf{82.0}\\
\hline
\end{tabular}
\label{Tab30}
\vspace*{-0.5em}
\end{table*}

\subsection{Evaluation on Occluded Person Datasets}
\textbf{Image Setting.}
Tab.~\ref{Tab1} summarizes the results of our method and previous works on image occluded dataset.  We also list the network backbone, loss function and extra information (key points and foreground) used by each method. For fair comparison, we re-implement RFCnet which uses the same backbone, loss function and extra information with the compared methods. In particular, when the extra key points are not used, RFCnet replaces APU by \textit{fixed} partition with 6 regions (shown in Fig.~\ref{partition}). When the extra foreground information are not used,  RFCnet removes the \textit{foreground-guided region feature extractor} of RFC blocks.

As shown in Tab.~\ref{Tab1}, RFCnet consistently achieves the best performance under the same conditions. It is noted that: \textbf{(1)} The gaps between our results and those methods designed for holistic reID~\cite{PCB,part-aligned,Fd-gan} are significant: about  $10\%$ improvement in term of mAP. This is because occlusions largely affect the feature extracting of holistic reID methods, leading to poor performance. \textbf{(2)} Some occluded reID methods~\cite{miao2019pose} only extract the non-occluded parts for relieving the influence of occlusions. RFCnet outperforms these methods with an improvement up to  $8.8\%$ on mAP. We argue that the completion strategy in RFCnet is conducive to distinguish the different identities with seeming alike visible parts. In addition, different from the existing methods that only solve occlusion in the end, RFCnet can alleviate the occlusion corruptions in the earlier stages in a progressive manner. \textbf{(3)} Some reID methods~\cite{zhong2017random,adversarially} employs a data augmentation mechanism which adds occluded images to the training set. RFCnet still achieve much better performance: about  $10\%$ improvement on mAP. Further, as a data augmentation technique, these works~\cite{zhong2017random,adversarially} are compatible with our method to further lift the performance.

\textbf{Video Setting.}
Tab.~\ref{Tab2} summarizes the result of our method and previous works on video occluded dataset. To make a fair comparison on Occluded-DukeMTMC-VideoReID, we implement several recent works using ResNet50 as backbone, which are trained with the combination of cross-entropy and triplet loss. Our method outperforms the best existing methods. \textbf{(1)} The works~\cite{RCN,Triplet} designed for holistic video reID treat each frame of a video equally, leading to the corruption of the video representation by occluded frames. RFCnet surpasses these works up to  $29.6\%$ and $27.7\%$ in term of top-1 accuracy and mAP. \textbf{(2)} RFCnet outperforms the occluded reID methods~\cite{diversity,QAN,RQAN} up to $15.2\%$ mAP. These methods leverage a temporal attention network to select the non-occluded frames to alleviate occlusion. But it still leads to the spatial information loss and temporal information interrupt. The significant improvements can be attributed to the spatial-temporal information enhancement by feature completion in RFC blocks. \textbf{(3)} Existing completion-based reID methods~\cite{VRSTC} employ an \textit{image} completion network to recover the appearance of occluded image pixels, which makes the reID framework too complicated and time consuming. Our RFCnet puts much less overhead with a much better performance: about  $13.4\%$ mAP gain. We argue that the improvement is due to our feature completion strategy where reID tasks can provide direct feedback to the completion task. 

\subsection{Evaluation on Holistic Person Datasets}
\begin{table*}[t]
\small
\centering
\caption{Comparison with related methods on MARS and DukeMTMC-VideoReID datasets.}
\vspace*{-0.5em}
\label{mars}
\begin{center}
\begin{tabular}{l |c | c c | c c| c  c|c c }
\hline
\multirow{2}*{Methods} & \multirow{2}*{Backbone}  & \multicolumn{2}{c|}{ Loss Function} & \multicolumn{2}{c|}{External Clues}  & \multicolumn{2}{c|}{Mars}  &\multicolumn{2}{c}{Duke-VideoReID} \\   
\cline{3-10}
& &CE & Triplet & key-points & Foreground &mAP  &top-1 &mAP &top-1 \\
\hline
STAN~\cite{diversity} &ResNet50  &\checkmark &$\times$ &$\times$ &$\times$ &65.8  &82.3 &-    &-     \\
EUG \cite{dukereid} &ResNet50 &\checkmark &$\times$ &$\times$ &$\times$ & 67.4  &80.8 &78.3 &83.6 \\
M3D~\cite{M3D} &ResNet50 &\checkmark &$\times$ &$\times$ &$\times$ & 74.1  &84.4 &- &-\\ 
Snippet      \cite{snippet}  &ResNet50 &\checkmark &$\times$ &$\times$ &$\times$  &76.1 &86.3 &- &-  \\
TAFD \cite{zhao2019attribute} &ResNet50 & \checkmark &$\times$ &$\times$ &$\times$ &78.2 &87.0 &- &-\\
GLTP~\cite{GLTL}  &ResNet50 & \checkmark &$\times$ &$\times$ &$\times$ & 78.5 & 87.0 &93.7 &96.3 \\
VRSTC~\cite{VRSTC} &ResNet50 & \checkmark &$\times$&$\times$ &$\times$ &82.3 &88.5 &93.8 &95.0 \\
\textbf{RFCnet} & ResNet50 & \checkmark&$\times$ &$\times$ &$\times$ & \textbf{83.1} &\textbf{88.6} &\textbf{95.5} & \textbf{95.6} \\
\hline
COSAM~\cite{Co-segmentation} &ResNet50 & \checkmark &\checkmark &$\times$ &$\times $&79.9 &84.9 &93.7 &96.2\\
\textbf{RFCnet} &ResNet50 &\checkmark & \checkmark& $\times$ & $\times$ & \textbf{85.7} & \textbf{90.5} & \textbf{96.6} & \textbf{96.8} \\
\hline
\textbf{RFCnet}   &ResNet50 & \checkmark& \checkmark & \checkmark & \checkmark  & \textbf{86.3}   &\textbf{90.7} & \textbf{97.0} & \textbf{97.6} \\
\hline
\end{tabular}
\end{center}
\label{tab3}
\vspace*{-0.8em}
\end{table*}

\textbf{Image Setting.}
Tab.~\ref{Tab30} compares RFCnet with the state-of-the-arts on four image reID benchmarks. Form the table, it can be seen that the proposed RFCnet achieves the competitive performance under all conditions. Firstly, RFCnet outperforms the occluded reID methods~\cite{miao2019pose,adversarially,sun2019perceive,he2019foreground}. RFCnet increases  $2.0\%$ and $2.2\%$ mAP on Market-1501 and DukeMTMC respectively. The improvements demonstrate that the proposed feature completion is also more effective on holistic reID task. Secondly, RFCnet achieves comparable even superior results to the methods designed specifically for holistic reID~\cite{PCB,IANet,zhang2019densely,zheng2019pyramidal}, which indicates the good generality of our method. On the other hand, our method only deals with the occlusion problem. 
 Thirdly, RFCnet significantly outperforms previous methods on CUHK03-NP and MSMT17 datasets, which shows the superiority of our method on more challenge scenes.
The effectiveness of our method also indicates that occlusion is an important issue and has important research significance in reID.

\textbf{Video Setting.} We then compare our method with state-of-the-arts on holistic video datasets. As shown in Tab.~\ref{tab3}, our method achieves the best performance. It significantly outperforms the temporal-attention based methods~\cite{diversity,GLTL} by about  $4.6\%$ mAP on MARS, further showing the effectiveness of completion strategy. In addition, RFCnet surpasses the holistic reID methods~\cite{M3D,snippet,zhao2019attribute} up to  $4\%$ mAP on MARS. We argue that the RFC blocks utilize the rich visible clues to recover the occluded information, which eliminates the impact of occlusion and enhances the feature representation. Compared to VRSTC~\cite{VRSTC}, RFCnet achieves $1.7\%$ mAP gains on DukeMTMC-VideoReID with less computations. The improvements further demonstrate the superiority of completion on the feature level.

Notably, compared to holistic reID datasets, our RFCnet shows more superiority on occluded datasets. It demonstrates that the existing reID methods perform poorly on occluded scenes, while our method can relieve influence of occlusion and achieve better performance.

\subsection{Ablation Study}
To investigate the effectiveness of each component in RFC block, we conduct a series of ablation studies on the image occluded dataset Occluded-DukeMTMC, and video occluded dataset Occluded-DukeMTMC-VideoReID. We adopt modified ResNet50 as the baseline. In particular, the baseline uses modified ResNet50 as the feature extractor, which is trained by the combination of cross entropy loss and triplet loss. For fair comparison, the training details of the baseline are the same with RFCnet.
Tab.~\ref{tab5} and \ref{tab6} summary the comparison results for different settings. If there is no special explanation, the proposed block is inserted to the last residual block of stage$_2$ layer of modified ResNet50.

\begin{table}[t]
\caption{Ablation study on image occluded reID task. Param: the parameter number of the models; GFLOPs: the number of floating-point operations for an input image;  TT: the training time of the models; IT: the inference time of retrieving all queries, where ``m'' denotes ``minutes'' and ``s'' denotes  ``seconds'' (The timings are performed on a server with one NVIDIA 2080Ti GPU).}
\centering
\small
\begin{tabular}{l |c c| c c | c c} 
\hline
\multirow{2}*{Models}  & \multicolumn{6}{c}{ Occluded-DukeMTMC} \\   
\cline{2-7}
&Param. &GFLOPs &  TT&  IT &mAP &top-1 \\
\hline
baseline &23.5M  & 4.06 & 72m & 110s  &45.8 &53.9  \\
Foreground &23.5M &4.06 & 73m & 110s  &47.1 &55.9\\
\hline
 Keypoint-Select &23.7M &4.18 &74m &111s &  48.1 &56.7\\
SRFC &23.9M &4.20 & 75m &112s  & \textbf{51.8}  & \textbf{60.9} \\
SRFC-A &- &- &- &- & 50.6  & 58.6\\
SRFC-P &- &- &- &- & 51.3   &60.0  \\
\hline
RFC (stage$_1$) &23.7M &4.21 &76m & 112s & 49.9 &58.9\\
RFC (stage$_2$) &23.9M &4.20 & 75m &112s  & 51.8  & 60.9\\
RFC (stage$_3$) &25.2M &4.20 &74m &111s  & \textbf{52.8} & \textbf{61.9}\\
RFC (stage$_4$) &30.0M &4.63 &80m &115s & 49.5 & 58.2\\
\hline
RFCnet-wo-$L_k$  &- &- &- &-  & 52.9 & 62.0 \\
RFCnet-wo-$L_f$  &- &- &- &-  & 52.5 & 61.4 \\
\hline
RFCnet-wo-$L_{ap}$ &- &- &- &- & 53.8&62.2 \\
RFCnet-wo-$L_a$  &- &- &- &-  & 54.0&62.5 \\
RFCnet-wo-$L_p$  &- &- &- &-  & 53.9 & 63.0 \\
RFCnet &25.6M &4.34 &78m &119s  & \textbf{54.5} & \textbf{63.9} \\
\hline
\end{tabular}
\label{tab5}
\vspace*{-1.2em}
\end{table}

\begin{table}[t]
\caption{Ablation study on video occluded reID task. Param: the parameter number of the models; GFLOPs: the number of floating-point operations for four-frames sequence;  TT: the training time of the models; IT: the inference time of retrieving all queries (The timings are performed on a server with two NVIDIA 2080Ti GPUs).}
\centering
\small
\begin{tabular}{L{2.105cm} | C{0.5cm} C{0.95cm}| C{0.5cm} C{0.5cm} |C{0.4cm}  C{0.7cm}} 
\hline
\multirow{2}*{Models}  &  \multicolumn{6}{c}{Occluded-DukeMTMC-VideoReID} \\   
\cline{2-7}
&Param. &GFLOPs &TT &IT&mAP &top-1 \\
\hline
baseline &23.5M  &16.24 &250m & 25m &69.3 &68.9  \\
Foreground &23.5M &16.25 &255m &25m &74.1 &73.2\\
\hline
 Keypoint-Select &23.7M &16.72 &264m &26m &  77.9 &76.7\\
SRFC &24.0M &16.81 &270m &26m& \textbf{82.9} & \textbf{82.6} \\
SRFC-A &- &-  &- &- & 80.6 & 79.1\\
SRFC-P &- &-  &- &- & 82.4  & 81.9  \\
\hline
TRFC &23.8M &16.80 &270m &27m &82.2 &82.2\\
S-T-RFC &24.2M &17.35 &280m &28m  & \textbf{89.4} &  \textbf{89.4}\\
T-S-RFC &- &- &- &- &88.9 &89.0\\
S+T-RFC &- &- &- &- &86.5 &  86.1\\
\hline
RFC (stage$_2$)  &24.2M &17.35 &280m &28m & 89.4 & 89.4\\
RFC (stage$_3$) &26.2M &17.36 &281m &29m & \textbf{91.1} & \textbf{91.5}\\
\hline
RFCnet-wo-$L_k$ &- &- &- &- &91.1& 92.9 \\
RFCnet-wo-$L_f$  &- &- &- &- & 90.8 & 91.1 \\
\hline
RFCnet-wo-$L_{ap}$ &- &- &- &- & 90.6 & 91.5 \\
RFCnet-wo-$L_a$ &- &- &- &- & 91.6 & 92.4 \\
RFCnet-wo-$L_p$  &- &- &- &- & 91.7 & 92.4 \\
\hline
 SRFCnet &-&- &- &- & 83.1 & 82.7\\
RFCnet &26.9M &18.47 &290m &31m & \textbf{92.0} &\textbf{93.0} \\
\hline
\end{tabular}
\label{tab6}
\vspace*{-1.2em}
\end{table}

\textbf{SRFC block\footnote{The SRFC block is formed by removing the TRFC module in RFC block.}.}
As shown in Tab.~\ref{tab5} and \ref{tab6}, SRFC block consistently improves the performance remarkably. Only employing the foreground masks (\textit{Foreground}) to discard the occluded regions only leads to small improvements. While SRFC significantly outperforms \textit{Foreground} by $4.7\%$ and $8.8\%$ mAP on Occluded-DukeMTMC and Occluded-DukeMTMC-VideoReID datasets respectively. \textbf{The significant improvements indicate that the completion operation to recover occluded regions is more effective than the previous discard strategy on occluded reID.}

We then investigate whether both appearance and position information are effective in SRFC. Tab.~\ref{tab5} and \ref{tab6} compares different SRFC blocks:  SRFC-A and  SRFC-P that respectively uses the appearance assignment matrix ($S^A$) and position assignment matrix ($S^P$) in the encoding process. As seen, only employing the appearance information (SRFC-A) brings  $4.8\%$ and  $11.3\%$ mAP  improvement on  Occluded-DukeMTMC and Occluded-DukeMTMC-VideoReID datasets respectively. We argue that the appearance information is conducive to complete the partially occluded region by using remaining visible appearance feature in this region. SRFC-P also achieves better results than baseline which demonstrates the effectiveness of introducing position prior. By combining the appearance and position information, SRFC consistently achieves the best performance. The results validate the complementary of appearance and position information for spatial completion.

\textbf{Compare SRFC to fixed selecting regions based on human keypoints.} In order to verify the effectiveness of the adaptive clustering in SRFC block, we further explore the strategy of grouping the regions based on the human keypoints (denoted as \textit{\textbf{keypoint-select}}). In particular, we respectively assign the regions belonging to ``head'' ($\{R^1\}$), ``upper-body'' ($\{R^2\}$) and  ``lower-body'' ($\{R^3,R^4,R^5,R^6\}$) to a cluster. As shown in Tab. \ref{tab5} and \ref{tab6}, SRFC significantly outperforms \textit{keypoint-select} by $4.2\%$ and $5.0\%$ mAP on Occluded-DukeMTMC and Occluded-DukeMTMC-VideoReID respectively. We argue that the fixed clustering of \textit{keypoint-select} is difficult to deal with all occlusion patterns, resulting in the performance degeneration. 

\textbf{TRFC block\footnote{The TRFC block is formed by removing the SRFC module in RFC block.}.} We further assess the effectiveness of TRFC block on video occluded reID task. As shown in Tab.~\ref{tab6} , TRFC individually outperforms baseline by above $10\%$ mAP and top-1 accuracy. We attribute the significant improvements to the effectiveness of capturing long-term temporal clues. TRFC blocks can attend the information from distant frames, which makes the completed features more precise and semantically consistent with the whole video sequence. When we integrate SRFC and TRFC blocks together to RFC block, the performance can be further lifted by $7.2\%$ mAP and top-1 accuracy. The reason is that SRFC and TRFC complete occluded features in different ways and will facilitate each other.

\textbf{Arrangement of SRFC and TRFC modules.} We compare three different ways of arranging SRFC and TRFC modules: sequential SRFC and TRFC blocks (S-T-RFC), sequential TRFC and SRFC blocks (T-S-RFC) and parallel use of both modules (S+T-RFC). As each module has different functions, the combining mode and order may affect the overall performance. Tab.~\ref{tab6} summarizes the experimental results on different arranging modes. We can see that the  sequentially arrangement performs better than the parallel mode. The sequential modes enable a progressive completion where one module gives a coarse prediction and the other refines the initial prediction. In this way, the completion difficulty is decomposed thus can produce a finer feature.  Note that all the combining methods outperform adding SRFC and TRFC independently, showing that utilizing both modules is crucial and the best-arranging strategy further pushes performance.

We observe that the spatial-first order performs slightly better than the temporal-first order.  We argue that S-T-RFC is more conducive to recover the  appearance of fully occluded regions. In particular, SRFC and TRFC use complementary contexts, where SRFC focuses on ``spatial'' with ``\textit{appearance and position}'' information, while TRFC focuses on ``temporal'' with ``\textit{only appearance}'' information.  So for a fully occluded region without appearance clues, it is difficult for TRFC to find the correlated frames, which may produce inaccurate feature recovering. On the contrary, SRFC can use  extra \textit{position} information to roughly recover the feature of the fully occluded region. After that, TRFC can use the recovered  appearance information to accurately attend the correlated non-occluded frame.  Therefore, we sequentially arrange SRFC and TRFC modules in this work. 

\begin{figure}[t]
\centering
   \includegraphics[width=1.0\linewidth]{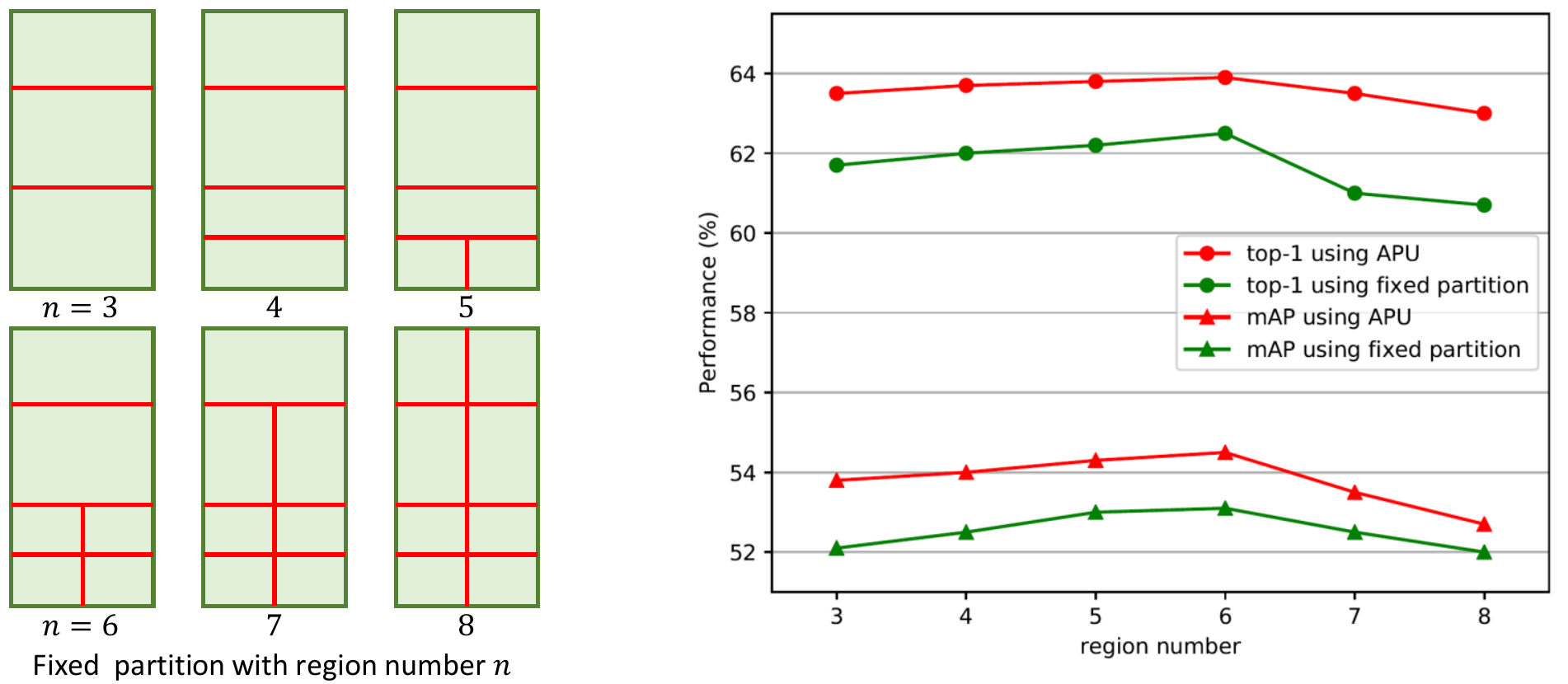}
   \caption{The top-1 accuracy and mAP using \textit{fixed} partition and \textit{adaptive} partition strategies (APU) with different region numbers on Occluded-DukeMTMC dataset.}
\label{partition}
\vspace*{-1em}
\end{figure}

\textbf{Location of RFC block within the network.} The effect of the RFC block is evaluated by plugging it after each stage of ResNet50. The network is trained and evaluated on Occluded-DukeMTMC dataset and the quantitative results are shown in Tab.~\ref{tab5}. From the results, it can be inferred that the inclusion of RFC block improves the baseline and is more effective on stage$_2$ and stage$_3$. We argue that the shallow feature is not very expressive and sufficient to provide precise semantic clues. In addition, the visual concepts in the last stage tend to be too abstract thus is difficult to propagate the spatial clues. In addition, the gains of adding RFC blocks on the middle stages over last stage verify the superiority of processing occlusion during feature extraction. 

Tab.~\ref{tab5} and Tab.~\ref{tab6} also show the results of inclusion of multiple RFC blocks simultaneously. As seen, RFC blocks can consistently lift the accuracy when more blocks are added. For example, on image occluded dataset, the model with RFC blocks plugged to stage$_2$ and stage$_3$ improve the model with one RFC block added to stage$_2$ by about $2.7\%$ mAP and $3.0\%$ top-1 accuracy. We argue that multiple RFC blocks form a hierarchical architecture, in which the second block can complete the occluded feature on the basic of the prediction of the first block and then provide some complementary features. 

\textbf{Adaptive Partition Strategy.} In this part, we evaluate the effectiveness of the proposed APU. As shown in Fig.~\ref{partition}, we evaluate multiple variants of our method that use fixed partition$/$APU \textit{w.r.t.} different region numbers. As seen, the proposed adaptive partition consistently outperforms the fixed partition, showing the superiority of adaptive partition strategy. We argue that the fixed partition is prone to the spatial misalignment of person images. Thus the regions produced by fixed partition lack precise semantics which is not conduct to the spatial-temporal correlation modeling for feature completion. 

We observe that APU achieves the best performance when the region number is $6$. We argue that: 1) too small region number cannot accurately locate the occluded body parts; 2) too large region number may divide the non-occluded part into multiple regions, which damages the semantics of the corresponding part. Specifically,  we observe that the occlusion typically occurs on the lower part of the pedestrians, and the head and upper-body parts are usually fully visible on Occluded-DukeMTMC and Occluded-DukeMTMC-VideoReID datasets. So when the divided region number is set to 8, the head and upper-body parts are both divided into two regions, which damages the semantics of the two parts, resulting in inferior performance.

APU is conceptually similar to STN~\cite{jaderberg2015spatial} because both are designed to learn a transformation matrix for informative part location. However, APU incorporates the human structure information, which utilizes the human keypoints to guide the learning of transformation. Without the keypoints guidance, STN easily produces a degenerate solution, where multiple transforms learn to detect the same body region. We further compare APU with STN where APU is replaced by STN in RFC block. STN achieves $49.1\%$ mAP and $53.9\%$ top-1. So APU significantly outperforms STN by $6.4\%$ mAP, which verifies the superiority of APU.

\begin{figure}[t]
\centering
   \includegraphics[width=1.0\linewidth]{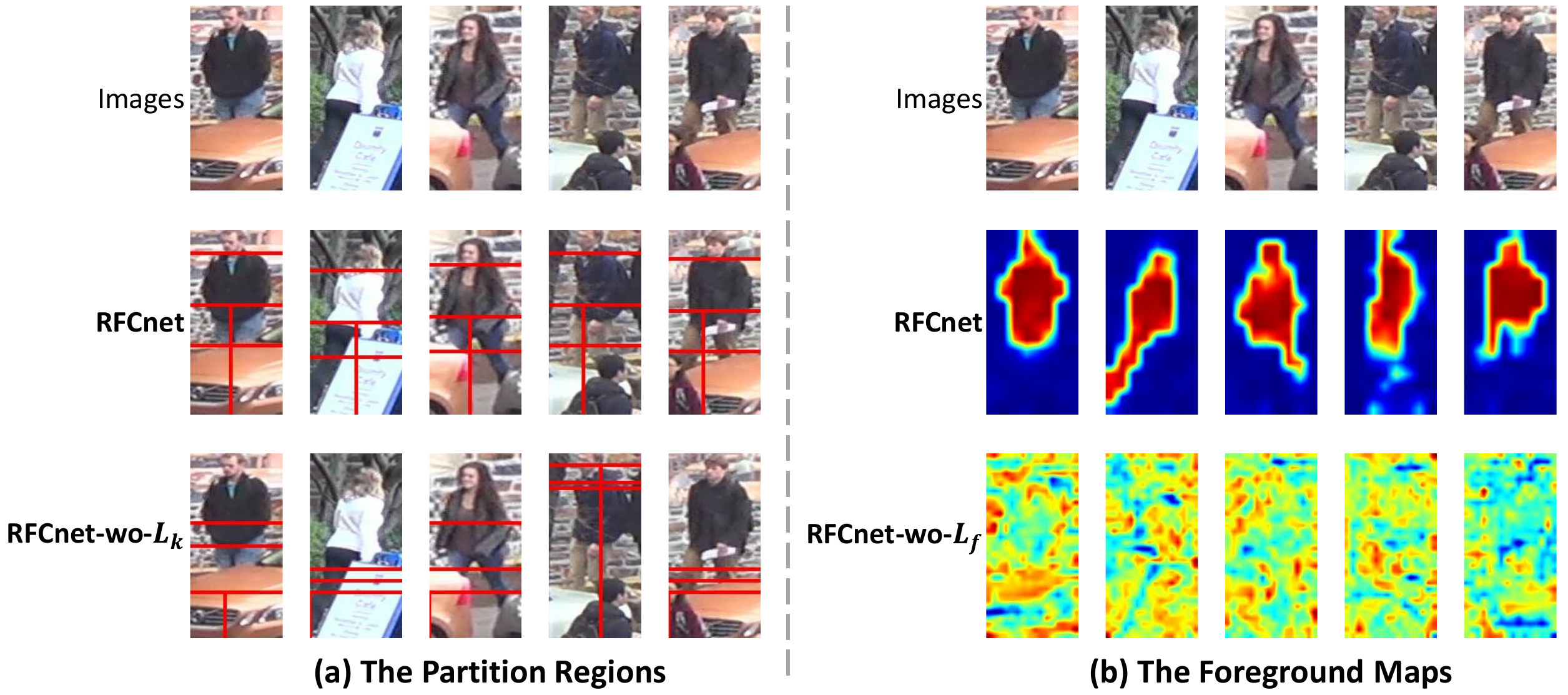}
   \caption{(a) Learned Partition Regions. Images and corresponding APU's partition results of RFCnet and RFCnet trained without key-points constraint $L_k$. (b) Learned Foreground Maps. Images and corresponding foreground maps generated by RFCnet and RFCnet trained without foreground maps constraint $L_f$.}
\label{vis-param}
\vspace*{-1em}
\end{figure}

\textbf{Effect of Key-points Constraint $L_k$.} To evaluate the contribution of proposed key-points constraint $L_k$, we train RFCnet without $L_k$ (RFCnet-wo-$L_k$) and report the results. As shown in Tab.~\ref{tab5} and \ref{tab6}, the results of RFCnet consistently outperform that of RFCnet-wo-$L_k$ on both image and video occluded benchmarks. This confirms the effectiveness of using key-points constraint. We also visualize the learnt partitions of RFCnet and RFCnet-wo-$L_k$ in Fig.~\ref{vis-param} (a). We observe that RFCnet is able to produce an adaptive meaningful partition, in which each region focuses on a specific body part. On the contrary, the divided regions tend to be disorganized and lack specific semantics in RFCnet-wo-$L_k$. Therefore, it is difficult for RFCnet-wo-$L_k$ to capture spatial-temporal clues between different body parts, resulting in a poor completion and performance degradation.

\begin{figure}[t]
\centering
   \includegraphics[width=0.9\linewidth]{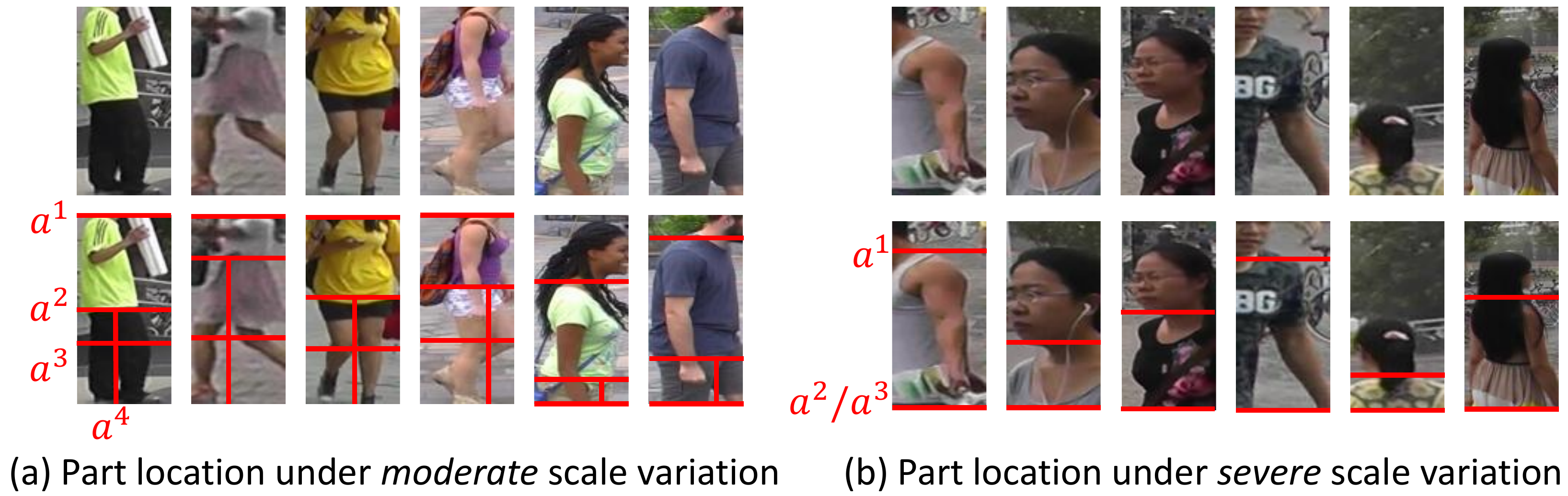}
   \caption{The partition results of APU in the scenario where input person images only contain a few body parts.}
\label{apu}
\vspace*{-1em}
\end{figure}

Notably, our keypoints locator can deal with the hard cases, where the person images only contain a few body parts due to imperfect pedestrian detection. Fig.~\ref{apu} visualizes some examples. We observe that: (1) the keypoints locator can still accurately locate the contained body parts even under severe scale variation. We argue that the locator aims to predict four \textbf{coarse} keypoints, which is less sensitive to scale variation.  (2) the locator learns to generate corresponding boundary values when the keypoints cannot be detected. For example, for the images in Fig.~\ref{apu} (b), the keypoints \textit{hip} and \textit{knee} are missed and the locator predicts both $a^2$ and $a^3$ to $h$. In this way, these images are divided into two regions, \textsl{i.e.}, head and upper body, which can generate accurate local representation of contained regions and empty representation of non-contained regions.

\textbf{Effect of Foreground Maps Constraint $L_f$.} To evaluate the effectiveness of foreground maps constraint $L_f$, we also show the results of RFCnet trained without $L_f$ (RFCnet-wo-$L_f$). Experimental results are presented in Tab.~\ref{tab5} and \ref{tab6}. We can clearly observe that RFCnet significantly outperforms RFCnet-wo-$L_f$, which demonstrates the effectiveness of foreground map constraint. Similarly, we visualize the learnt foreground maps of RFCnet and RFCnet-wo-$L_f$ in Fig.~\ref{vis-param} (b). As seen, the generated foreground maps of RFCnet can accurately detect the person parts. On the contrary, the learned foreground maps of RFCnet-wo-$L_f$ tend to be messy and may focus on background and occluded regions. Therefore, it is difficult for RFCnet-wo-$L_f$ to eliminate the interference of obstacles, resulting in performance degradation. 

\begin{figure}[t]
\centering
   \includegraphics[width=0.7\linewidth]{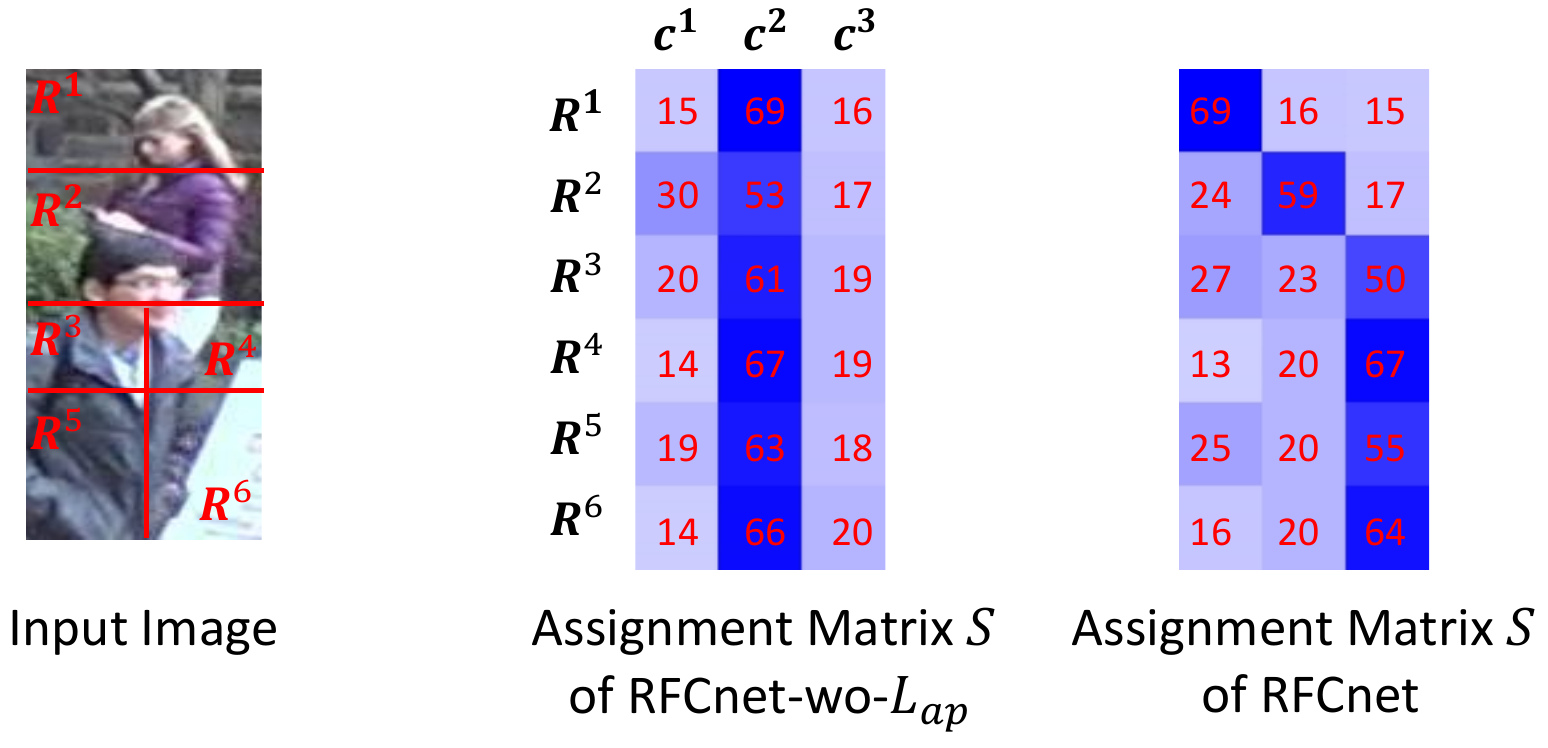}
   \caption{Learned Assignment Matrix $S$ of RFCnet-wo-$L_{ap}$ and RFCnet for input image.}
\label{vis-assign}
\vspace*{-1em}
\end{figure}

\textbf{Effect of Assignment Regularization Terms $L_a$ and $L_p$.} To evaluate the effectiveness of assignment regularizations, we respectively train RFCnet without $L_a+L_p/L_a/L_p/$ (RFCnet-wo-$L_{ap}$$/$RFCnet-wo-$L_a$$/$RFCnet-wo-$L_p$). As shown in Tab.~\ref{tab5} and \ref{tab6}, RFCnet consistently outperforms the other three models, indicating the effectiveness of proposed assignment regularizations.

We also visualize the learnt assignment matrix $S$ of RFCnet-wo-$L_{ap}$ and RFCnet in Fig.~\ref{vis-assign}. As seen, RFCnet-wo-$L_{ap}$ may produce degenerate assignment matrix for severely occluded image. We argue that when a person is severely occluded, there is little valid information to be used. In this case, RFCnet-wo-$L_{ap}$  is difficult to identify which regions are relevant, and tends to assign all regions to a cluster. On the contrary, the \textit{assignment regularization terms} ($L_a/L_p$) explicitly constrain the similarity of assignment vectors to be consistent with the appearance$/$position similarity of corresponding regions. Therefore, for the appearance-dissimilar regions ($R^1$ and $R^2$ in Fig.~\ref{vis-assign}) and distant regions ($R^1$ and $\{R^i\}_{i=3}^6$), RFCnet can learn to assign them to different clusters, producing a reasonable assignment matrix.

\textbf{Complexity Comparisons.} To illustrate the cost of RFC blocks, we report the number of network parameters (Params), the number of floating-point operations (GFLOPs) and training and testing time in Tab.~\ref{tab5} and Tab.~\ref{tab6}. We can observe that RFC block introduces small computational and time overhead. For example on image reID task, RFCnet requires $4.34$ GFLOPs, corresponding to only $6.8\%$ relative increase over original model. RFCnet introduces $2.1$M parameters corresponding to $8.9\%$ relative increase over baseline.
And RFCnet requires 78 minutes(m) training time and 119 seconds(s) inference time, corresponding to $8.3\%$ and $8.2\%$ relative increase over baseline. The small additional overhead required by RFCnet is justified by its contribution to model performance.

\subsection{Performance Analysis between video and image.}
\begin{figure}[t]
\centering
   \includegraphics[width=0.9\linewidth]{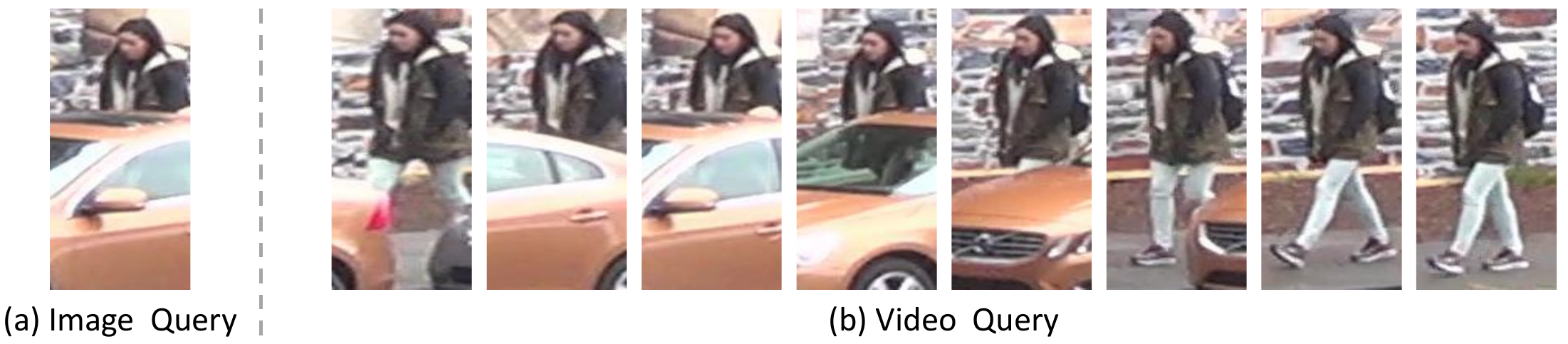}
   \vspace*{-0.5em}
   \caption{A query of the same pedestrian on Occluded-DukeMTMC and Occluded-DukeMTMC-VideoReID.}
\label{image}
\vspace*{-1em}
\end{figure}

As shown in Tab.~\ref{Tab1} and \ref{Tab2}, RFCnet achieves much better performance on Occluded-DukeMTMC-VideoReID than Occluded-DukeMTMC datasets. We argue that there are two main reasons for this phenomenon. \textbf{Firstly,  the occlusion issue can be partially solved by the non-occluded frames of input sequence on Occluded-DukeMTMC-VideoReID}. Fig.~\ref{image} visualizes a query of the same pedestrian  on  Occluded-DukeMTMC and Occluded-DukeMTMC-VideoReID. On Occluded-DukeMTMC, the lower-body of the person is completely occluded, so the model is only able to distinguish the identity based on upper-body part. While the integral appearance of the person can be obtained by the non-occluded frames on Occluded-DukeMTMC-VideoReID, which alleviates the information loss caused by occlusion. Also, as shown in Tab.~\ref{tab5} and \ref{tab6}, baseline achieves $15\%$ higher performance on  Occluded-DukeMTMC-VideoReID than Occluded-DukeMTMC, which implicitly shows that the input sequences with more frames can partially solve the occlusion issue.

\textbf{Secondly, the proposed \textit{Temporal Region Feature Completion} (TRFC) block brings additional performance gains on Occluded-DukeMTMC-VideoReID}. As shown in Fig.~\ref{image}, the lower-body of the person is completely occluded with no appearance clues. In this case, it is difficult for SRFC block to accurately predict its feature.  On the contrary, TRFC can use the visible information from the non-occluded frames, which is conducive to accurately recover the appearance of occluded frames thus achieves better performance. In addition, as shown in Tab.~\ref{tab5} and \ref{tab6}, SRFC brings $10\%$ and $13.8\%$ top-1 gains over baseline on Occluded-DukeMTMC and Occluded-DukeMTMC-VideoReID respectively, and TRFC brings $10.3\%$ more top-1 gain on Occluded-DukeMTMC-VideoReID. The results also show that the additional performance gain over baseline on Occluded-DukeMTMC-VideoReID is mainly brought by TRFC block.

\subsection{Parameter Analysis}
\begin{figure}[t]
\centering
   \includegraphics[width=1.0\linewidth]{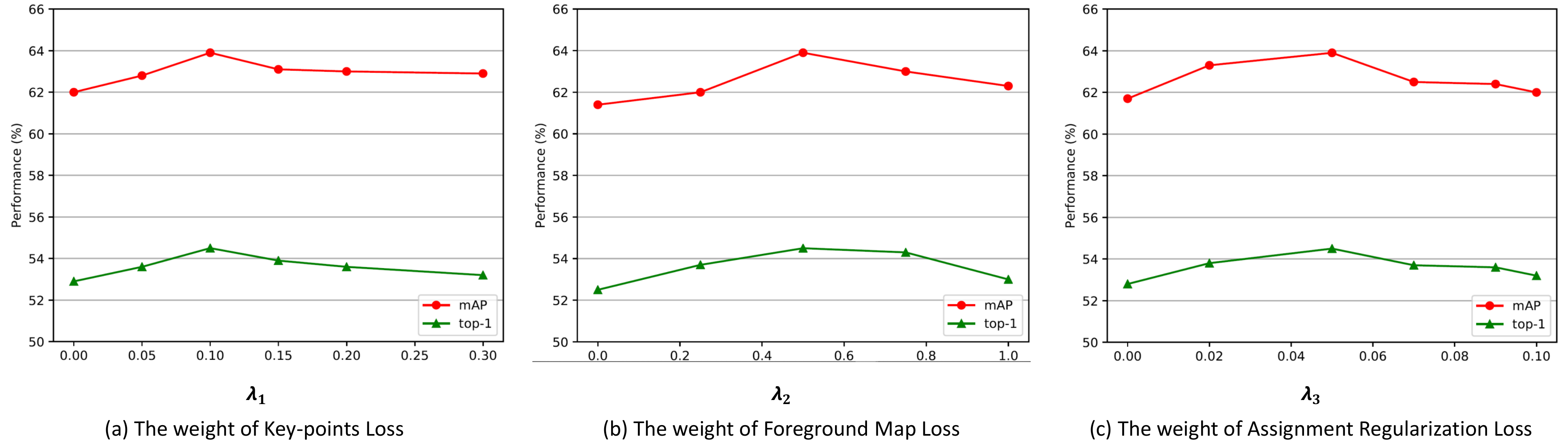}
   \caption{ The top-1 accuracy and mAP on Occluded-DukeMTMC dataset using (a) different $\lambda_1$, fixed $\lambda_2=0.5$ and $\lambda_3=0.05$. (b) different $\lambda_2$, fixed $\lambda_1=0.1$ and $\lambda_3=0.05$. (c) different $\lambda_3$, fixed $\lambda_1=0.1$ and $\lambda_2=0.5$.}
\label{param}
\vspace*{-1em}
\end{figure}

 Fig.~\ref{param} evaluates the influence of the hyper-parameters $\lambda_1$, $\lambda_2$ and $\lambda_3$ (Eq.~\ref{eq17}) on Occluded-DukeMTMC dataset. We respectively evaluate each hyper-parameter, where we change its value and fix the other hyper-parameters to the optimal values. Notably, we split 100 identities from the original training set as the validation set, and use the validation set to tune the hyper-parameters. The phenomenon \textit{w.r.t} different hyper-parameters on validation set is consistent with that on test set. So we only show the performance on the test set in Fig.~\ref{param}.

\textit{Firstly}, we change $\lambda_1$ from $0$ to $0.3$ to learn different models. As shown in Fig.~\ref{param} (a), the performance of different $\lambda_1>0$ consistently outperforms that of $\lambda_1=0$, which further verifies the effect of key-points constraint $L_k$. In addition, the performance of different $\lambda_1$ is stable, indicating the robustness of RFCnet to various $\lambda_1$. We can find that RFCnet achieves the best performance when $\lambda_1=0.1$.  \textit{Secondly}, we change $\lambda_2$ from 0 to 1.0. As shown in Fig.~\ref{param} (b), the performance of different $\lambda_2>0$ fluctuates in a small range and achieves the best performance when $\lambda_2=0.5$. \textit{Finally}, we change $\lambda_3$ from 0 to 0.1. We observe that the performance of different $\lambda_3>0$ consistently outperforms that of $\lambda_3=0$, indicating the effect of assignment regularization terms. And RFCnet achieves the best performance when $\lambda_3=0.05$. Therefore, we choose to use $\lambda_1=0.1$, $\lambda_2=0.5$ and $\lambda_3=0.05$ in our work.

\subsection{Visualization RFC Block}
\begin{figure}[t]
\centering
   \includegraphics[width=0.9\linewidth]{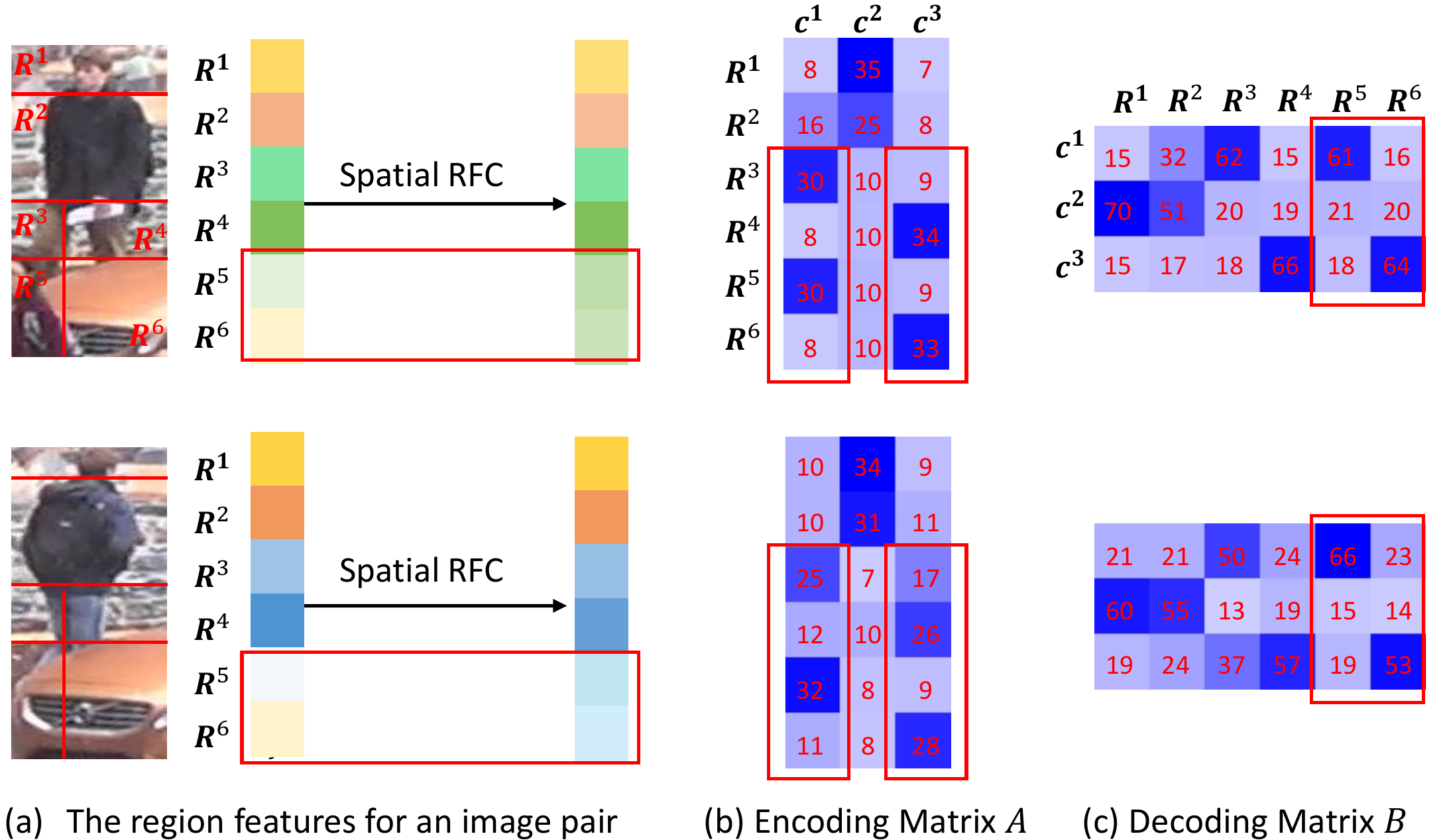}
   \caption{Completed features of SRFC module. (a) Visualization of the initial region feature $f$ and updated region feature $o$ by SRFC module for input image pair. The dimensionality of $f$ and $o$ is reduced to $N\times1$ by PCA for visualization. (b) The encoding matrix of SRFC module ($A$). For the values $x$ in $A$, we present $100x$ for better visualization, and a darker color indicates a higher value. (c) The decoding matrix of SRFC module ($B$).}
\label{vis-SRFC}
\vspace*{-1em}
\end{figure}

\textbf{Visualizing SRFC module.}
For qualitative analysis, we visualize the completed features by SRFC, where we insert a RFC block after stage$_3$ layer of ResNet50. 
Fig.~\ref{vis-SRFC} visualizes the initial region features ($f$ in Fig.~\ref{SFC}), the completed region features by SRFC module ($o$ in Fig.~\ref{SFC}), the encoding matrix ($A$) and decoding matrix ($B$) respectively. It is clear that, for the two persons occluded by a car, the appearance information of the lower-leg almost disappears in the initial feature $f$. In addition, since the two persons wear similar upper clothes, the features of upper body are not very discriminative. The above two factors make the initial features insufficient to distinguish the two pedestrians. On the contrary, SRFC module completes the features of lower-body, making the updated features more distinguishable. As seen, $R^5$ is partially occluded  where the remaining visible area has a similar appearance to $R^3$. With the proposed \textit{appearance assignment regularization}, the encoding matrix assigns $R^5$ and $R^3$ to a cluster $c^1$. And the fully occluded region $R^6$ is very close to $R^4$. With the proposed  \textit{position assignment regularization}, the encoding matrix assigns $R^4$ and $R^6$ to a cluster $c^3$. Then the decoding matrix distributes the feature of $c^1/c^3$ to $R^5/R^6$. So the occluded region can use the information of correlated non-occluded region in the cluster to recover its appearance.

\begin{figure}[t]
\centering
   \includegraphics[width=0.7\linewidth]{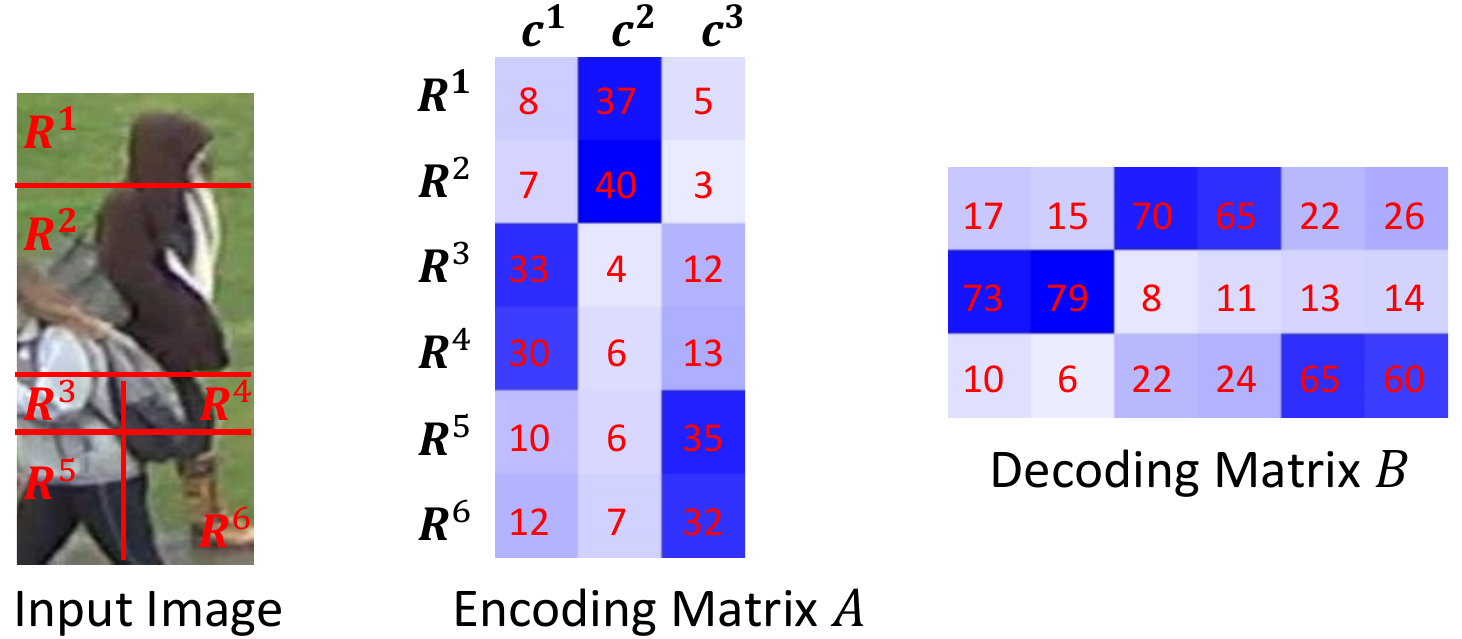}
   \caption{Examples of encoding matrix $A$ and decoding matrix $B$ for input images.}
\label{vis-SRFC1}
\vspace*{-1em}
\end{figure}

Notably, the assignment of SRFC is adaptive to the occlusion mode of input image. For example, as shown in Fig.~\ref{vis-SRFC1}, the regions $R^3$ and $R^5$ of the input image are fully occluded. With the \textit{appearance assignment regularization}, SRFC assigns appearance-dissimilar regions $R^4$ (pants) and $R^6$ (shoes) to different clusters. With the  \textit{position assignment regularization}, SRFC assigns $R^3$ and the closest $R^4$ to a cluster $c^1$, and $R^5$ and the closest $R^6$ to a cluster $c^3$. Overall, SRFC produces a assignment mode that is different from Fig.~\ref{vis-SRFC}, which can still effectively recover the features of occluded regions.

\begin{figure}[t]
\centering
   \includegraphics[width=0.9\linewidth]{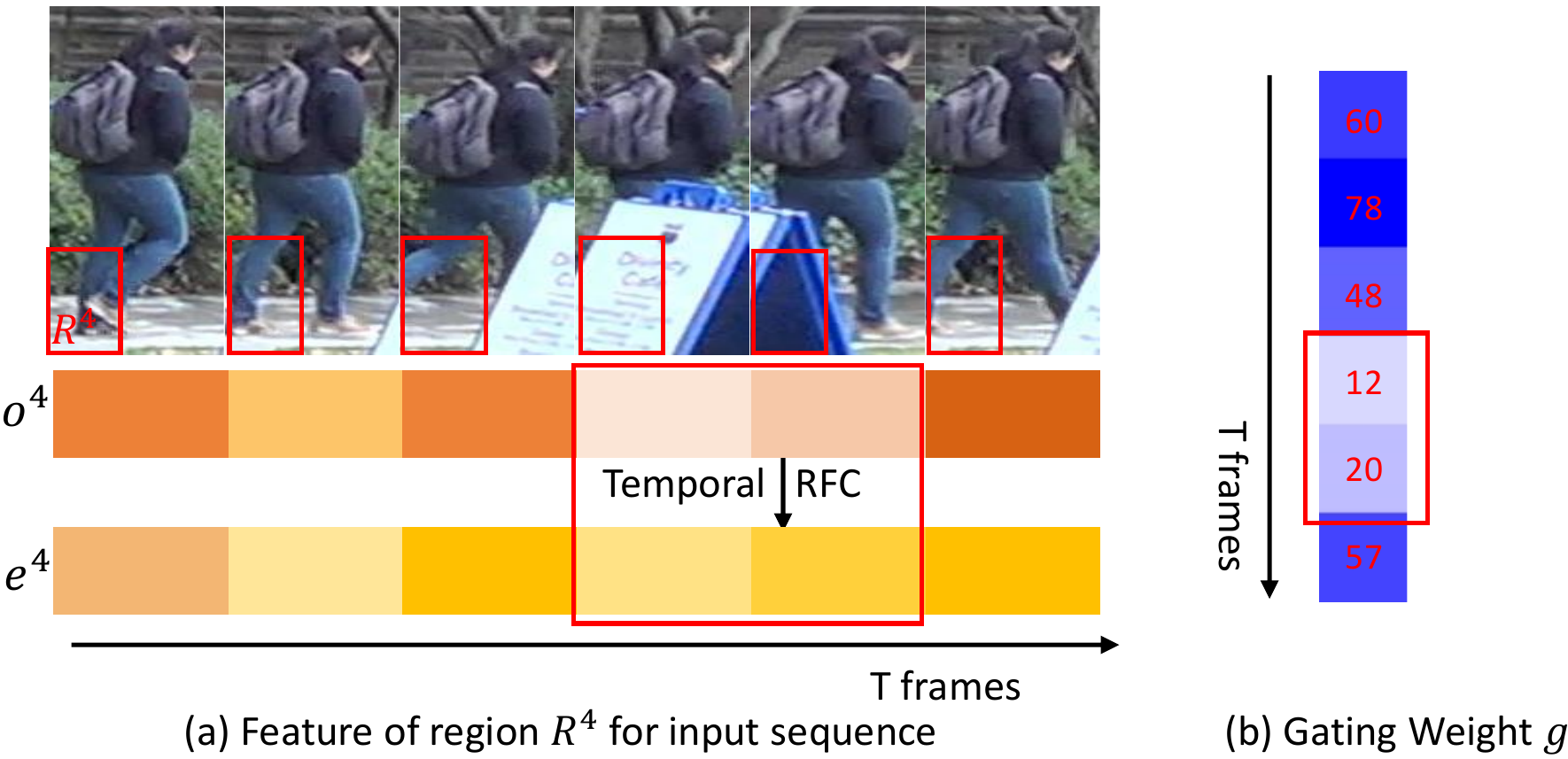}
   \caption{Completed features of TRFC module. (a) Visualization of the initial feature $o^4$ of region $R^4$ and updated feature $e^4$ by TRFC module for input sequence. The dimensionality of $o^4$ and $e^4$ is reduced to $T\times1$ by PCA for visualization. (b) The gating weight of TRFC module ($g$).}
\label{vis-TRFC}
\vspace*{-1em}
\end{figure}

\textbf{Visualizing TRFC module.}
We further visualize the completed features by TRFC, where we insert a RFC block after stage$_3$ layer of ResNet50.  Fig.~\ref{vis-TRFC} visualizes the initial features of region $R^4$ ($o^4$ in Fig.~\ref{TFC}), updated region feature ($e^4$ in Fig.~\ref{TFC}) and the gating weight ($g$ in Fig.~\ref{TFC}) of the input sequence. It is clear that, the occlusion affects the initial region features, \textit{i.e.,} the feature substantially changes as occlusion happens. TRFC can utilize the memory mechanism to capture the temporal clues of other frames.  Therefore, with the aggregation of other frames' information, the features of occluded regions can be recovered to describe the original body parts, as shown in Fig.~\ref{vis-TRFC} (b). We can also observe that TRFC assigns lower gating weights to the occluded regions. It indicates that the features of occluded regions are suppressed during temporal completion operation and the final video feature can better describe the target person. 

\subsection{Visualizing Retrieval Results}
\begin{figure}[t]
\centering
   \includegraphics[width=0.8\linewidth]{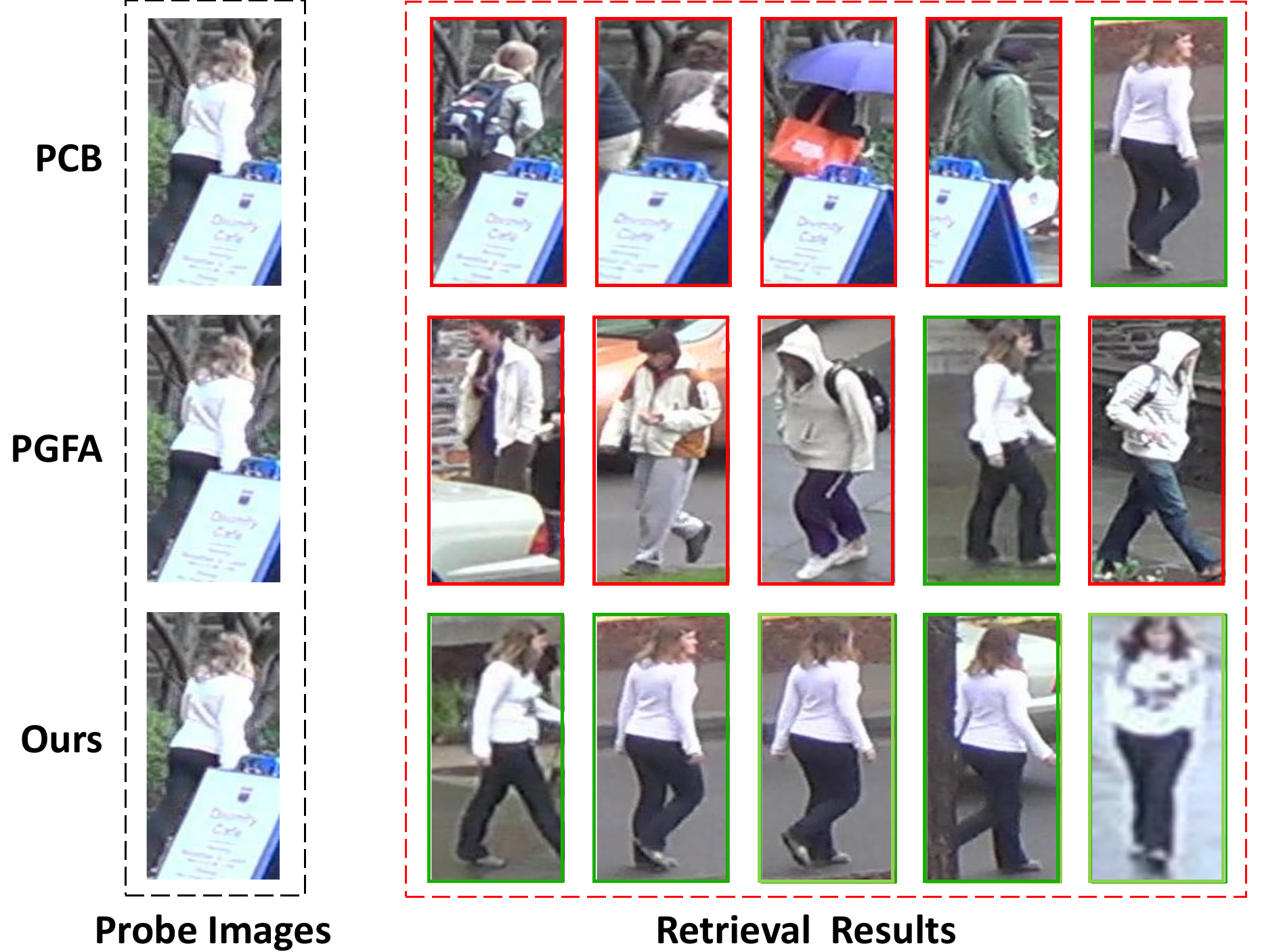}
   \caption{Comparison of PCB~\cite{PCB}, PGFA~\cite{miao2019pose} and our RFCnet. Green and red rectangles indicate correct and error retrieval results, respectively.}
\label{vis-Retrie}
\vspace*{-1em}
\end{figure}

In this section, we qualitatively demonstrate the performance of RFCnet by showing some ranking examples. Fig.~\ref{vis-Retrie} shows some retrieval examples of existing methods, PCB~\cite{PCB} and PGFA~\cite{miao2019pose}, and our RFCnet method on Occluded-DukeMTMC. The retrieval results show that PCB is prone to mix the information of the target person and obstacles, resulting in retrieving a wrong person with similar obstacle. Although  PGFA utilizes the pose landmarks to alleviate the interference of obstacle, it still exists the information loss. As shown in Fig.~\ref{vis-Retrie}, because of discarding the occluded regions, the characteristic of upper-body dominates the probe feature in PGFA. This makes PGFA tend to retrieve the wrong pedestrians with similar upper clothes. On the contrary, our RFCnet can complete the lower-body feature and form a full characteristic of the target identity. So RFCnet can work successfully in the case.

\begin{figure}[t]
\centering
   \includegraphics[width=0.9\linewidth]{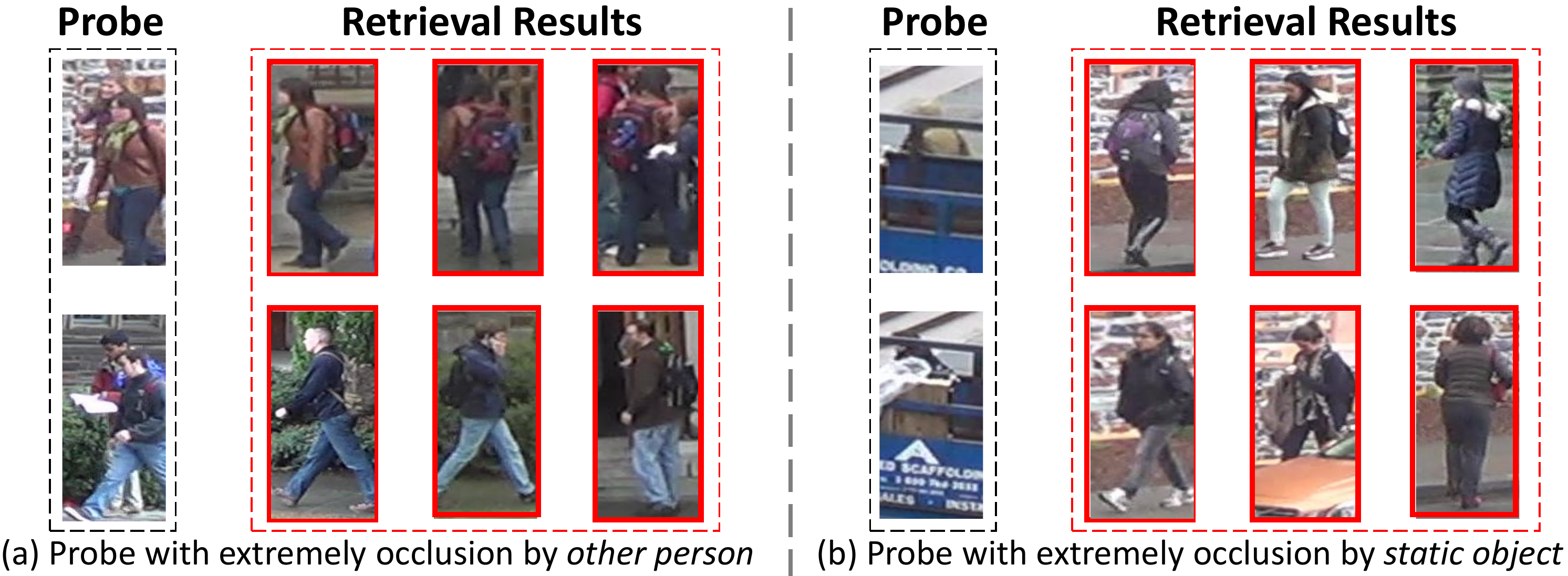}
   \caption{Illustration of failure cases by our method. The red rectangles indicate error retrieval results, respectively.}
\label{limit}
\vspace*{-1.2em}
\end{figure}

\textbf{Failure Cases.} We additionally show some failure cases of RFCnet in Fig.~\ref{limit}. Our method may fail to deal with the cases with severe occlusion. (1) When a person is severely occluded by other person, the model may mistake the disturbed person as the target pedestrian, resulting in wrong retrieval results. For example, as illustrated in Fig.~\ref{limit} (a), given a probe image where the target person is severely occluded by another pedestrian, our model retrieves wrong person images with similar appearance to the disturbed person. (2) When a person is extremely occluded by static object, it is difficult for our model to predict the features of occluded parts since most body parts are completely invisible. In this case, our model can just utilize extremely insufficient appearance information, which may fail to retrieve the true person images. Some examples can be seen in Fig.~\ref{limit} (b). 

\section{Conclusion}
In this paper, we propose a RFC block for occluded person reID. The RFC block jointly captures the long-range spatial context and long-term temporal contexts for recovering the occluded regions. By recovering the occluded features, our method can suppress the noise and alleviate the information loss caused by occluded regions on the target person. Besides, to facilitate the research on the Video Occluded reID problem, we introduce a large-scale dataset, Occluded-DukeMTMC-VideoReID. Extensive experiments on image and video reID demonstrate the effectiveness of our proposed method, especially on occluded datasets. 

Occluded person reID remains largely an unsolved problem and we argue that: (1) The completion strategy that recovers the appearance of occluded regions usually performs better than the discarding strategy to ignore the occluded regions. The discarding strategy only considers the non-occluded parts to measure the partial similarity, leading to measure inconsistency. While the completion strategy can obtain a complete representation of the target person, which is conducive to distinguish the persons with similar non-occluded parts. (2) The feature level completion is superior to image level completion. The feature completion can capture the longer-range spatial and longer-term temporal contexts to achieve better completion. Moreover, the feature completion is lightweight and can be readily inserted into a reID network end-to-end. So we suggest that the feature level completion strategy should be given priority for occluded reID task.

In the future, we intend to explore a better combination mode of spatial and temporal completion mechanism for occluded reID, such as collaborative integration. We will also apply our method in other computer vision tasks on occlusion scenes, \textit{e.g.}, tracking and detection.



\ifCLASSOPTIONcompsoc
  \section*{Acknowledgments}
\else
  \section*{Acknowledgment}
\fi

This work is partially supported by National Key R$\&$D Program of China (No. 2017YFA0700800) and Natural Science Foundation of China (NSFC): 61876171 and 61976203, and the Open Project Fund from Shenzhen Institute of Artificial Intelligence and Robotics for Society, under Grant No. AC01202005015.

\ifCLASSOPTIONcaptionsoff
  \newpage
\fi

\bibliographystyle{ieeetr}
\bibliography{egbib}

\begin{thebibliography}{10}

\bibitem{zhuo2018occluded}
J.~Zhuo, Z.~Chen, J.~Lai, and G.~Wang, ``Occluded person re-identification,''
  in {\em IEEE International Conference on Multimedia and Expo}, pp.~1--6,
  2018.

\bibitem{miao2019pose}
J.~Miao, Y.~Wu, P.~Liu, Y.~Ding, and Y.~Yang, ``Pose-guided feature alignment
  for occluded person re-identification,'' in {\em IEEE International
  Conference on Computer Vision}, pp.~542--551, 2019.

\bibitem{sun2019perceive}
Y.~Sun, Q.~Xu, Y.~Li, C.~Zhang, Y.~Li, S.~Wang, and J.~Sun, ``Perceive where to
  focus: Learning visibility-aware part-level features for partial person
  re-identification,'' in {\em IEEE Conference on Computer Vision and Pattern
  Recognition}, pp.~393--402, 2019.

\bibitem{he2019foreground}
L.~He, Y.~Wang, W.~Liu, H.~Zhao, Z.~Sun, and J.~Feng, ``Foreground-aware
  pyramid reconstruction for alignment-free occluded person
  re-identification,'' in {\em IEEE International Conference on Computer
  Vision}, pp.~8450--8459, 2019.

\bibitem{VRSTC}
R.~Hou, B.~Ma, H.~Chang, X.~Gu, S.~Shan, and X.~Chen, ``V{RSTC}: Occlusion-free
  video person re-identification,'' in {\em IEEE Conference on Computer Vision
  and Pattern Recognition}, pp.~7183--7192, 2019.

\bibitem{context}
D.~Pathak, P.~Krahenbuhl, J.~Donahue, T.~Darrell, and A.~A. Efros., ``Context
  encoders: Feature learning by inpainting,'' in {\em IEEE Conference on
  Computer Vision and Pattern Recognition}, pp.~2536--2544, 2016.

\bibitem{non-local}
X.~Wang, R.~Girshick, A.~Gupta, and K.~He, ``Non-local neural networks,'' in
  {\em IEEE Conference on Computer Vision and Pattern Recognition},
  pp.~7794--7803, 2018.

\bibitem{residual}
K.~He, X.~Zhang, S.~Ren, and J.~Sun, ``Deep residual learning for image
  recognition,'' in {\em IEEE Conference on Computer Vision and Pattern
  Recognition}, pp.~770 -- 778, 2016.

\bibitem{dukereid}
Y.~Wu, Y.~Lin, X.~Dong, Y.~Yan, W.~Quyang, and Y.~Yang, ``Exploit the unknown
  gradually: One-shot video-based person re-identification by stepwise
  learning,'' in {\em IEEE Conference on Computer Vision and Pattern
  Recognition}, pp.~5177--5186, 2018.

\bibitem{background}
M.~Tian, S.~Yi, H.~Li, S.~Li, X.~Zhang, J.~Shi, J.~Yan, and X.~Wang,
  ``Eliminating background-bias for robust person re-identification,'' in {\em
  IEEE Conference on Computer Vision and Pattern Recognition}, pp.~5794--5803,
  2018.

\bibitem{smoothed}
S.~Bai, X.~Bai, and Q.~Tian, ``Scalable person re-identification on supervised
  smoothed manifold,'' in {\em IEEE Conference on Computer Vision and Pattern
  Recognition}, pp.~2530--2539, 2017.

\bibitem{Learning}
S.~Paisitkriangkrai, C.~Shen, and A.~van~den Hengel, ``Learning to rank in
  person re-identification with metric ensembles.,'' in {\em IEEE Conference on
  Computer Vision and Pattern Recognition}, pp.~1846--1855, 2015.

\bibitem{hard-aware}
R.~Yu, Z.~Dou, S.~Bai, Z.~Zhang, Y.~Xu, and X.~Bai, ``Hard-aware point-to-set
  deep metric for person re-identification,'' in {\em European Conference on
  Computer Vision}, pp.~188--204, 2018.

\bibitem{gu2019temporal}
X.~Gu, B.~Ma, H.~Chang, S.~Shan, and X.~Chen, ``Temporal knowledge propagation
  for image-to-video person re-identification,'' in {\em IEEE International
  Conference on Computer Vision}, pp.~9647--9656, 2019.

\bibitem{zhu2017fast}
X.~Zhu, B.~Wu, D.~Huang, and W.~Zheng, ``Fast open-world person
  re-identification,'' {\em IEEE Transactions on Image Processing}, vol.~27,
  no.~5, pp.~2286--2300, 2017.

\bibitem{li2019unsupervised}
M.~Li, X.~Zhu, and S.~Gong, ``Unsupervised tracklet person re-identification,''
  {\em IEEE Transactions on Pattern Analysis and Machine Intelligence}, 2019.

\bibitem{wang2016person}
T.~Wang, S.~Gong, X.~Zhu, and S.~Wang, ``Person re-identification by
  discriminative selection in video ranking,'' {\em IEEE Transactions on
  Pattern Analysis and Machine Intelligence}, vol.~38, no.~12, pp.~2501--2514,
  2016.

\bibitem{spindle-net}
H.~Zhao, M.~Tian, S.~Sun, J.~Shao, J.~Yan, S.~Yi, X.~Wang, and X.~Tang,
  ``Spindle net: Person re-identification with human body region guided feature
  decomposition and fusion,'' in {\em IEEE Conference on Computer Vision and
  Pattern Recognition}, pp.~1077--1085, 2017.

\bibitem{short-term}
R.~R. Varior, B.~Shuai, J.~Lu, D.~Xu, and G.~Wang, ``A siamese long short-term
  memory architecture for human reidentification.,'' in {\em European
  Conference on Computer Vision}, pp.~135--153, 2016.

\bibitem{point-to-set}
S.~Zhou, J.~Wang, J.~Wang, Y.~Gong, and N.~Zheng, ``Point to set similarity
  based deep feature learning for person re-identification,'' in {\em IEEE
  Conference on Computer Vision and Pattern Recognition}, pp.~5028 -- 5037,
  2017.

\bibitem{multi-channel}
D.~Cheng, Y.~Gong, S.~Zhou, J.~Wang, and N.~Zheng, ``Person re-identification
  by multi-channel parts-based cnn with improved triplet loss function,'' in
  {\em IEEE Conference on Computer Vision and Pattern Recognition}, pp.~1335 --
  1344, 2016.

\bibitem{KPM}
Y.~Shen, T.~Xiao, H.~Li, S.~Yi, and X.~Wang, ``End-to-end deep
  kronecker-product matching for person re-identification,'' in {\em IEEE
  Conference on Computer Vision and Pattern Recognition}, pp.~6886--6895, 2018.

\bibitem{MSL}
X.~Li, W.-S. Zheng, X.~Wang, T.~Xiang, and S.~Gong, ``Multi-scale learning for
  low-resolution person re-identification,'' in {\em IEEE International
  Conference on Computer Vision}, pp.~3765--3773, 2015.

\bibitem{Cuhk}
W.~Li, R.~Zhao, T.~Xiao, and X.~Wang, ``Deepreid: Deep filter pairing neural
  network for person re-identification.,'' in {\em IEEE Conference on Computer
  Vision and Pattern Recognition}, pp.~152--159, 2014.

\bibitem{triplet-loss}
S.~Ding, L.~Lin, G.~Wang, and H.~Chao, ``Deep feature learning with relative
  distance comparison for person re-identification,'' {\em Pattern
  Recognition}, 2015.

\bibitem{mars}
L.~Zheng, Z.~Bie, Y.~Sun, J.~Wang, C.~Su, S.~Wang, and Q.~Tian, ``Mars: A video
  benchmark for large-scale person re-identification,'' in {\em European
  Conference on Computer Vision}, pp.~868--884, 2016.

\bibitem{zhang2019densely}
Z.~Zhang, C.~Lan, W.~Zeng, and Z.~Chen, ``Densely semantically aligned person
  re-identification,'' in {\em IEEE Conference on Computer Vision and Pattern
  Recognition}, pp.~667--676, 2019.

\bibitem{mask-guided}
C.~Song, Y.~Huang, W.~Ouyang, and L.~Wang, ``Mask-guided contrastive attention
  model for person reidentification,'' in {\em IEEE Conference on Computer
  Vision and Pattern Recognition}, pp.~1179--1188, 2018.

\bibitem{semantic}
M.~M. Kalayeh, E.~Basaran, M.~Gökmen, M.~E. Kamasak, and M.~Shah, ``Human
  semantic parsing for person re-identification,'' in {\em IEEE Conference on
  Computer Vision and Pattern Recognition}, pp.~1062--1071, 2018.

\bibitem{pose-invariant}
L.~Zheng, Y.~Huang, H.~Lu, and Y.~Yang, ``Pose-invariant embedding for deep
  person re-identification,'' {\em IEEE Transactions on Image Processing},
  vol.~28, no.~9, pp.~4500--4509, 2019.

\bibitem{pose-driven}
C.~Su, J.~Li, S.~Zhang, J.~Xing, W.~Gao, and Q.~Tian, ``Pose-driven deep
  convolutional model for person re-identification,'' in {\em IEEE
  International Conference on Computer Vision}, pp.~3960--3969, 2017.

\bibitem{guo2019beyond}
J.~Guo, Y.~Yuan, L.~Huang, C.~Zhang, J.-G. Yao, and K.~Han, ``Beyond human
  parts: Dual part-aligned representations for person re-identification,'' in
  {\em IEEE International Conference on Computer Vision}, pp.~3642--3651, 2019.

\bibitem{zheng2015partial}
W.-S. Zheng, X.~Li, T.~Xiang, S.~Liao, J.~Lai, and S.~Gong, ``Partial person
  re-identification,'' in {\em IEEE Conference on Computer Vision and Pattern
  Recognition}, pp.~4678--4686, 2015.

\bibitem{TDL}
J.~You, A.~Wu, X.~Li, and W.~Zheng, ``Top-push video-based person
  re-identification,'' in {\em IEEE Conference on Computer Vision and Pattern
  Recognition}, pp.~1345--1353, 2016.

\bibitem{K-liu}
K.~Liu, B.~Ma, W.~Zhang, and R.~Huang, ``A spatiotemporal appearance
  representation for video-based pedestrian re-identification,'' in {\em IEEE
  International Conference on Computer Vision}, pp.~3810--3818, 2015.

\bibitem{stepwise}
Z.~Liu, D.~Wang, and H.~Lu, ``Stepwise metric promotion for unsupervised video
  person re-identification,'' in {\em IEEE International Conference on Computer
  Vision}, pp.~2429--2438, 2017.

\bibitem{ilids}
T.~Wang, S.~Gong, X.~Zhu, and S.~Wang, ``Person reidentification by video
  ranking,'' in {\em European Conference on Computer Vision}, pp.~688--703,
  2014.

\bibitem{RCN}
N.~McLaughlin, J.~M. del Rincon, and P.~C. Miller, ``Recurrent convolutional
  network for video-based person re-identification,'' in {\em IEEE Conference
  on Computer Vision and Pattern Recognition}, pp.~1325--1334, 2016.

\bibitem{jointly}
S.~Xu, Y.~Cheng, K.~Gu, Y.~Yang, S.~Chang, and P.~Zhou, ``Jointly attentive
  spatial-temporal pooling networks for video-based person re-identification,''
  in {\em IEEE International Conference on Computer Vision}, pp.~4743--4752,
  2017.

\bibitem{See}
Z.~Zhou, Y.~Huang, W.~Wang, L.~Wang, and T.~Tan., ``See the forest for the
  trees: Joint spatial and temporal recurrent neural networks for video-based
  person re-identification,'' in {\em IEEE Conference on Computer Vision and
  Pattern Recognition}, pp.~6776--6785, 2017.

\bibitem{snippet}
D.~Chen, H.~Li, T.~Xiao, S.~Yi, and X.~Wang, ``Video person re-identification
  with competitive snippet-similarity aggregation and co-attentive snippet
  embedding,'' in {\em IEEE Conference on Computer Vision and Pattern
  Recognition}, pp.~1169--1178, 2018.

\bibitem{V3DP}
X.~Liao, L.~He, and Z.~Yang, ``Video-based person re-identification via 3d
  convolutional networks and non-local attention,'' in {\em Asian Conference on
  Computer Vision}, pp.~620--634, 2018.

\bibitem{GLTL}
J.~Li, J.~Wang, Q.~Tian, W.~Gao, and S.~Zhang, ``Global-local temporal
  representations for video person re-identification,'' in {\em IEEE
  International Conference on Computer Vision}, 2019.

\bibitem{QAN}
Y.~Liu, J.~Yan, and W.~Ouyang, ``Quality aware network for set to set
  recognition,'' in {\em IEEE Conference on Computer Vision and Pattern
  Recognition}, pp.~4694--4703, 2017.

\bibitem{diversity}
S.~Li, S.~Bak, P.~Carr, C.~Hetang, and X.~Wang., ``Diversity regularized
  spatiotemporal attention for video-based person re-identification,'' in {\em
  IEEE Conference on Computer Vision and Pattern Recognition}, pp.~369--378,
  2018.

\bibitem{RQAN}
G.~Song, B.~Leng, Y.~Liu, C.~Hetang, and S.~Cai, ``Region-based quality
  estimation network for large-scale person re-identification,'' in {\em AAAI
  Conference on Artificial Intelligence}, vol.~32, 2018.

\bibitem{PCB}
Y.~Sun, L.~Zheng, Y.~Yang, Q.~Tian, and S.~Wang, ``Beyond part models: Person
  retrieval with refined part pooling (and a strong convolutional baseline),''
  in {\em European Conference on Computer Vision}, pp.~480--496, 2018.

\bibitem{PNGAN}
X.~Qian, Y.~Fu, W.~Wang, T.~Xiang, Y.~Wu, Y.~G. Jiang, and X.~Xue,
  ``Pose-normalized image generation for person re-identification.,'' in {\em
  European Conference on Computer Vision}, pp.~650--667, 2018.

\bibitem{liang2018look}
X.~Liang, K.~Gong, X.~Shen, and L.~Lin, ``Look into person: Joint body parsing
  \& pose estimation network and a new benchmark,'' {\em IEEE Transactions on
  Pattern Analysis and Machine Intelligence}, vol.~41, no.~4, pp.~871--885,
  2018.

\bibitem{part-aligned}
L.~Zhao, X.~Li, J.~Wang, and Y.~Zhuang, ``Deeply-learned part-aligned
  representations for person re-identification,'' in {\em IEEE International
  Conference on Computer Vision}, pp.~3239 -- 3248, 2017.

\bibitem{harmoniou}
W.~Li, X.~Zhu, and S.~Gong, ``Harmonious attention network for person
  re-identification,'' in {\em IEEE Conference on Computer Vision and Pattern
  Recognition}, pp.~2285 -- 2294, 2018.

\bibitem{attention-aware}
J.~Xu, R.~Zhao, F.~Zhu, H.~Wang, and W.~Quyang, ``Attention-aware compositional
  network for person re-identification,'' in {\em IEEE Conference on Computer
  Vision and Pattern Recognition}, pp.~2119--2128, 2018.

\bibitem{self-attention}
H.~Zhang, I.~Goodfellow, D.~Metaxas, and A.~Odena, ``Self-attention generative
  adversarial networks,'' in {\em International Conference on Machine
  Learning}, pp.~7354--7363, 2019.

\bibitem{BN}
S.~Ioffe and C.~Szegedy, ``Batch normalization: Accelerating deep network
  training by reducing internal covariate shift,'' in {\em International
  Conference on Machine Learning}, pp.~448--456, 2015.

\bibitem{imagenet}
A.~Karpathy, G.~Toderici, S.~Shetty, T.~Leung, R.~Sukthankar, and L.~Fei-Fei,
  ``Large-scale video classification with convolutional neural networks,'' in
  {\em IEEE Conference on Computer Vision and Pattern Recognition},
  pp.~1725--1732, 2014.

\bibitem{Triplet}
A.~Hermans, L.~Beyer, and B.~Leibe, ``In defense of the triplet loss for person
  reidentification,'' {\em arXiv preprint arXiv: 1703.07737}, 2017.

\bibitem{Ling-biometric}
L.~He, Z.~Sun, Y.~Zhu, and Y.~Wang., ``Recognizing partial biometric
  patterns.,'' {\em arXiv preprint arXiv:1810.07399}, 2018.

\bibitem{Suh-part-aligned}
Y.~Suh, J.~Wang, S.~Tang, T.~Mei, and K.~M. Lee, ``Part-aligned bilinear
  representations for person re-identification.,'' in {\em European Conference
  on Computer Vision}, pp.~402--419, 2018.

\bibitem{zhong2017random}
Z.~Zhong, L.~Zheng, G.~Kang, S.~Li, and Y.~Yang, ``Random erasing data
  augmentation,'' in {\em AAAI Conference on Artificial Intelligence}, vol.~34,
  pp.~13001--13008, 2020.

\bibitem{he2018deep}
L.~He, J.~Liang, H.~Li, and Z.~Sun, ``Deep spatial feature reconstruction for
  partial person re-identification: Alignment-free approach,'' in {\em IEEE
  Conference on Computer Vision and Pattern Recognition}, pp.~7073--7082, 2018.

\bibitem{adversarially}
H.~Huang, D.~Li, Z.~Zhang, X.~Chen, and K.~Huang, ``Adversarially occluded
  samples for person re-identification,'' in {\em IEEE Conference on Computer
  Vision and Pattern Recognition}, pp.~5098--5107, 2018.

\bibitem{Fd-gan}
Y.~Ge, Z.~Li, H.~Zhao, G.~Yin, S.~Yi, and X.~Wang, ``Fd-gan: Pose-guided
  feature distilling gan for robust person re-identification.,'' in {\em
  Advances in neural information processing systems}, pp.~1222--1233, 2018.

\bibitem{Market1501}
L.~Zheng, L.~Shen, L.~Tian, S.~Wang, J.~Wang, and Q.~Tian, ``Scalable person
  re-identification: A benchmark,'' in {\em IEEE International Conference on
  Computer Vision}, pp.~1116--1124, 2015.

\bibitem{Duke}
Z.~Zheng, L.~Zheng, and Y.~Yang, ``Unlabeled samples generated by gan improve
  the person re-identification baseline in vitro,'' in {\em IEEE International
  Conference on Computer Vision}, pp.~3754--3762, 2017.

\bibitem{msmt17}
L.~Wei, S.~Zhang, W.~Gao, and Q.~Tian, ``Person trasfer gan to bridge domain
  gap for person re-identification,'' in {\em Proc. IEEE Conference on Computer
  Vision and Pattern Recognition}, pp.~79--88, 2018.

\bibitem{multicamera}
E.~Ristani, F.~Solera, R.~Zou, R.~Cucchiara, and C.~Tomasi, ``Performance
  measures and a data set for multi-target, multicamera tracking,'' in {\em
  European Conference on Computer Vision}, pp.~17--35, 2016.

\bibitem{re-rank}
Z.~Zhong, L.~Zheng, D.~Cao, and S.~Li, ``Re-ranking person re-identification
  with k-reciprocal encoding.,'' in {\em IEEE Conference on Computer Vision and
  Pattern Recognition}, pp.~1318--1327, 2017.

\bibitem{DPM}
P.~Felzenszwalb, D.~McAllester, and D.~Ramanan, ``A discriminatively trained,
  multiscale, deformable part model,'' in {\em IEEE Conference on Computer
  Vision and Pattern Recognition}, pp.~1--8, 2008.

\bibitem{dehghan2015gmmcp}
A.~Dehghan, S.~Modiri~Assari, and M.~Shah, ``Gmmcp tracker: Globally optimal
  generalized maximum multi clique problem for multiple object tracking,'' in
  {\em IEEE Conference on Computer Vision and Pattern Recognition},
  pp.~4091--4099, 2015.

\bibitem{adam}
D.~P. Kingma and J.~Ba, ``Adam: A method for stochastic optimization,'' {\em
  arXiv preprint arXiv:1412.6980}, 2014.

\bibitem{zheng2019re}
M.~Zheng, S.~Karanam, Z.~Wu, and R.~J. Radke, ``Re-identification with
  consistent attentive siamese networks,'' in {\em IEEE Conference on Computer
  Vision and Pattern Recognition}, pp.~5735--5744, 2019.

\bibitem{IANet}
R.~Hou, B.~Ma, H.~Chang, X.~Gu, S.~Shan, and X.~Chen,
  ``Interaction-and-aggregation network for person re-identification,'' in {\em
  IEEE Conference on Computer Vision and Pattern Recognition}, pp.~9317--9326,
  2019.

\bibitem{dai2019batch}
Z.~Dai, M.~Chen, X.~Gu, S.~Zhu, and P.~Tan, ``Batch dropblock network for
  person re-identification and beyond,'' in {\em IEEE International Conference
  on Computer Vision}, pp.~3691--3701, 2019.

\bibitem{zheng2019pyramidal}
F.~Zheng, C.~Deng, X.~Sun, X.~Jiang, X.~Guo, Z.~Yu, F.~Huang, and R.~Ji,
  ``Pyramidal person re-identification via multi-loss dynamic training,'' in
  {\em IEEE Conference on Computer Vision and Pattern Recognition},
  pp.~8514--8522, 2019.

\bibitem{M3D}
J.~Li, S.~Zhang, and T.~Huang, ``Multiscale 3d convolution network for video
  based person reidentification,'' in {\em AAAI}, 2019.

\bibitem{zhao2019attribute}
Y.~Zhao, X.~Shen, Z.~Jin, H.~Lu, and X.-s. Hua, ``Attribute-driven feature
  disentangling and temporal aggregation for video person re-identification,''
  in {\em IEEE Conference on Computer Vision and Pattern Recognition},
  pp.~4913--4922, 2019.

\bibitem{Co-segmentation}
A.~Subramaniam, A.~Nambiar, and A.~Mittal, ``Co-segmentation inspired attention
  networks for video-based person re-identification,'' in {\em IEEE
  International Conference on Computer Vision}, pp.~562--572, 2019.

\bibitem{jaderberg2015spatial}
M.~Jaderberg, K.~Simonyan, A.~Zisserman, {\em et~al.}, ``Spatial transformer
  networks,'' in {\em Advances in neural information processing systems},
  pp.~2017--2025, 2015.

\end{thebibliography}

\begin{IEEEbiography}
[{\includegraphics[width=1in,height=1.25in,clip,keepaspectratio]{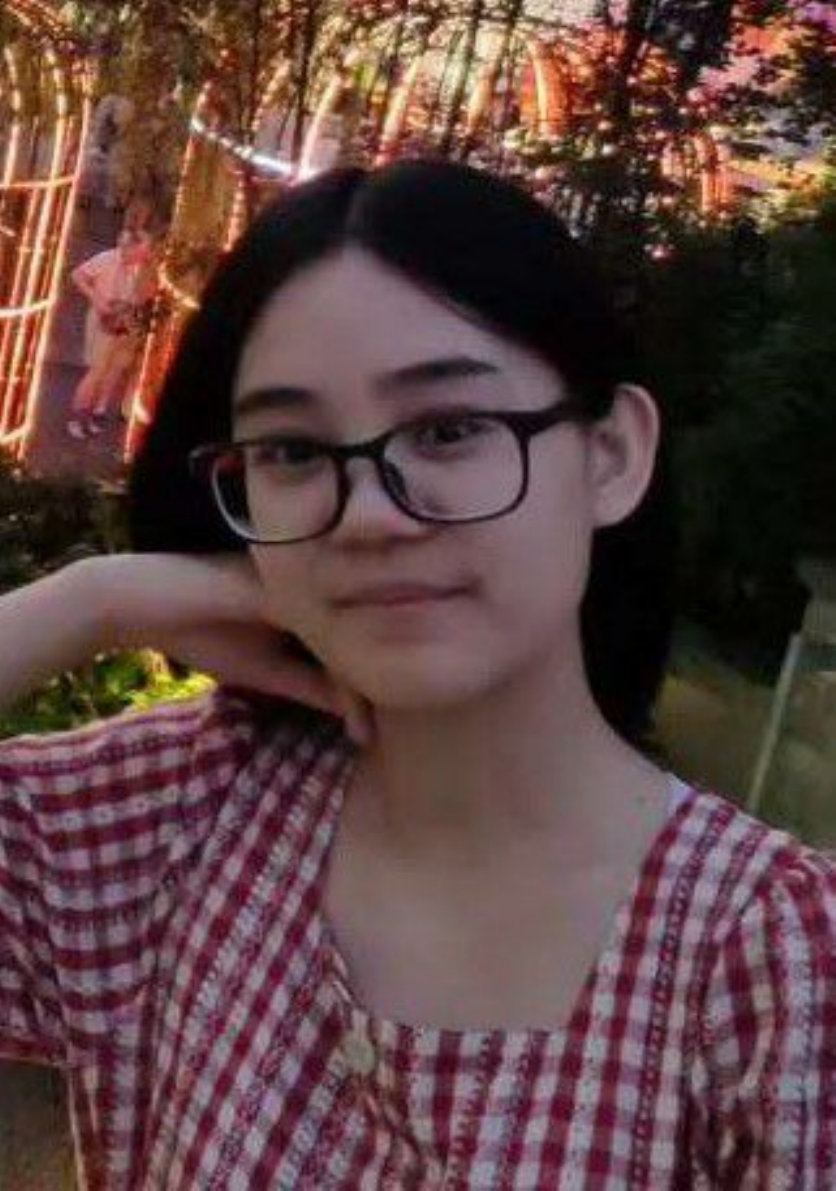}}]{Ruibing Hou}
received the BS degree in Northwestern Polytechnical University, Xi’an, China, in 2016. She is currently pursuing the Ph.D degree with the Institute of Computing Technology, Chinese Academy of Sciences, since 2016. Her research interests are in machine learning and computer vision. She specially focuses on person re-identification and few-shot learning. 
\end{IEEEbiography}

\begin{IEEEbiography}
[{\includegraphics[width=1in,height=1.25in,clip,keepaspectratio]{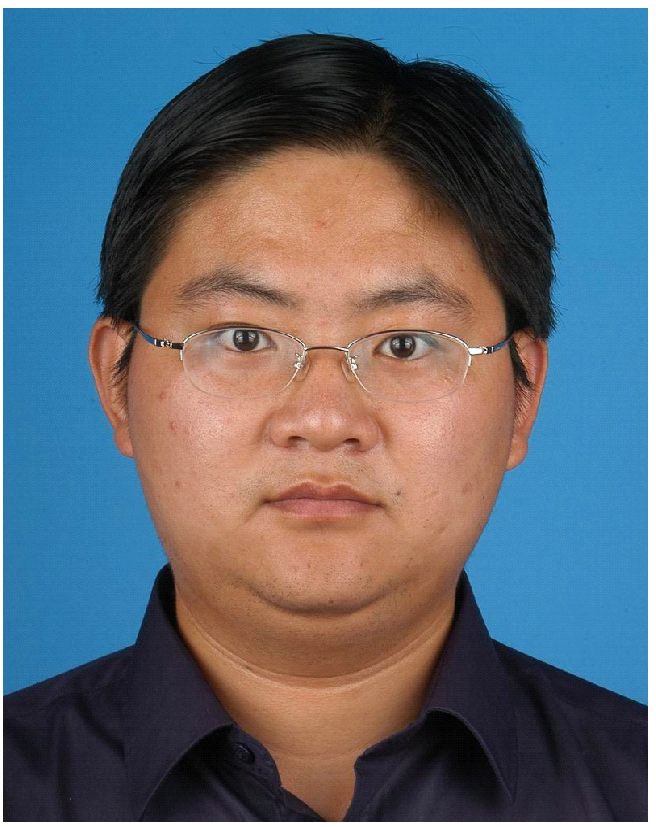}}]{Bingpeng Ma}
received the BS degree in mechanics, in 1998 and the MS degree in mathematics, in 2003 from the Huazhong University of Science and Technology, respectively. He received the PhD degree in computer science from the Institute of Computing Technology, Chinese Academy of Sciences, P.R. China, in 2009. He was a post-doctorial researcher with the University of Caen, France, from 2011 to 2012. He joined the School of Computer Science and Technology, University of Chinese Academy of Sciences, Beijing, in March 2013 and now he is an associate professor.  His research interests cover computer vision, pattern recognition, and machine learning. He especially focuses on person re-identification, face recognition, and the related research topics.
\end{IEEEbiography}

\begin{IEEEbiography}
[{\includegraphics[width=1in,height=1.25in,clip,keepaspectratio]{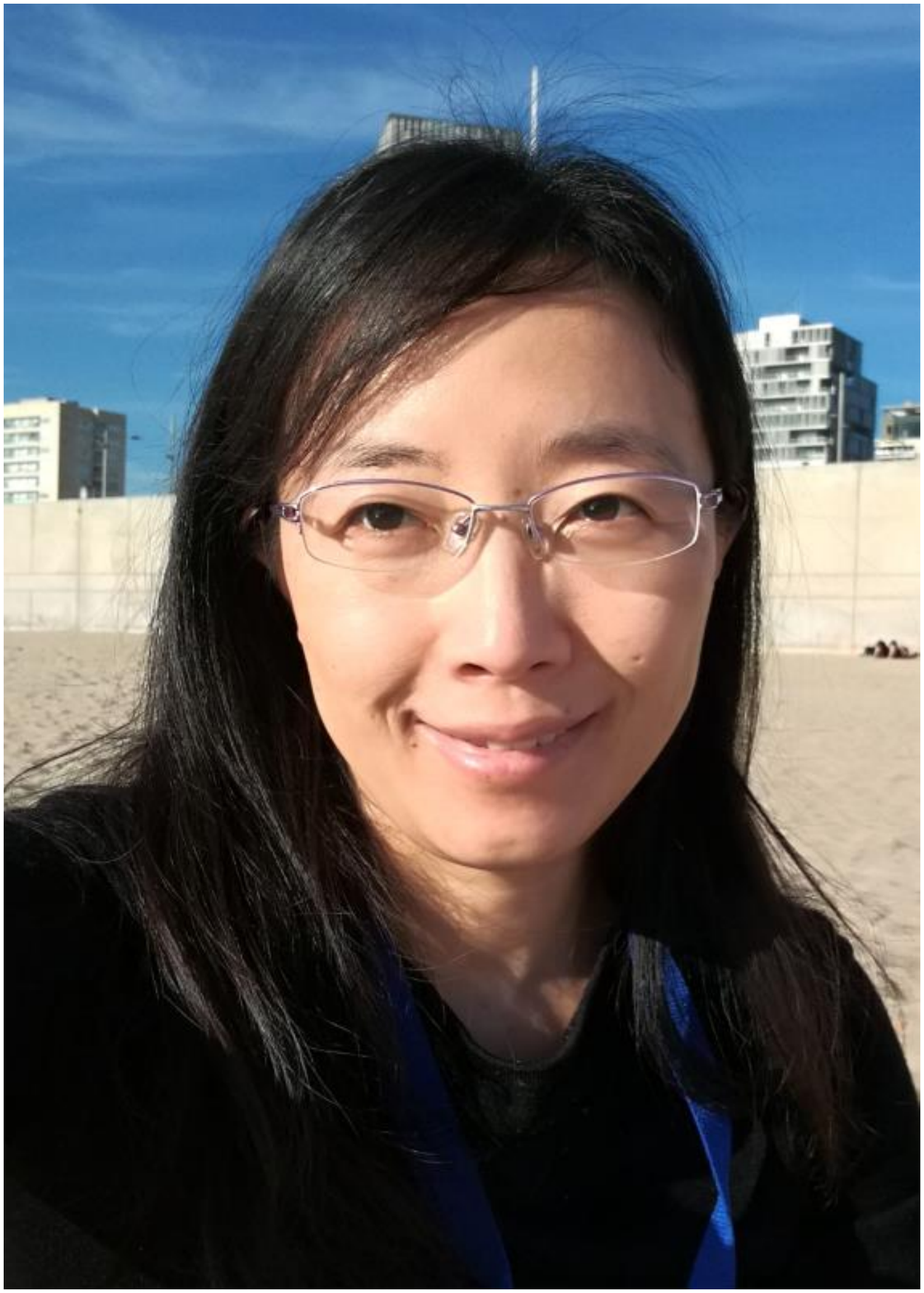}}]{Hong Chang}
received the Bachelor’s degree from Hebei University of Technology, Tianjin,
China, in 1998; the M.S. degree from Tianjin University, Tianjin, in 2001; and the Ph.D. degree from Hong Kong University of Science and Technology, Kowloon, Hong Kong, in 2006, all in computer science. She was a Research Scientist with Xerox Research Centre Europe. She is currently a Researcher with the Institute of Computing Technology, Chinese Academy of Sciences, Beijing, China. Her main research interests include algorithms and models in machine learning, and their applications in pattern recognition and computer vision.
\end{IEEEbiography}


\begin{IEEEbiography}
[{\includegraphics[width=1in,height=1.25in,clip,keepaspectratio]{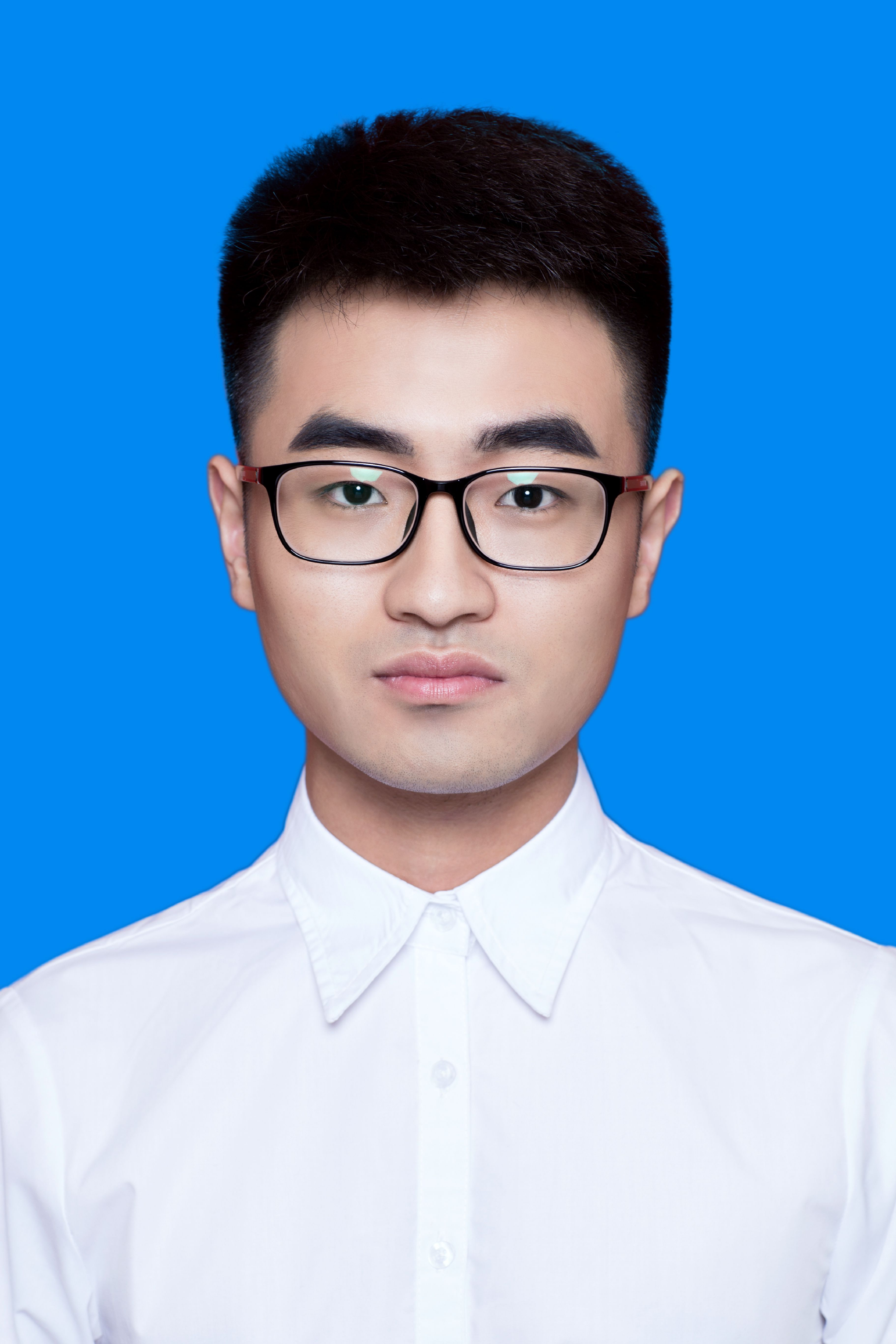}}]{Xinqian Gu}
received the BS degree in software engineering from Chongqing University in 2017. He is a PhD student at the Institute of Computing Technology (ICT), Chinese Academy of Sciences (CAS) as of 2017. His research interests are in computer vision, pattern recognition, and machine learning. He especially focuses on person re-identification, video analytics and the related research topics.
\end{IEEEbiography}


\begin{IEEEbiography}
[{\includegraphics[width=1in,height=1.25in,clip,keepaspectratio]{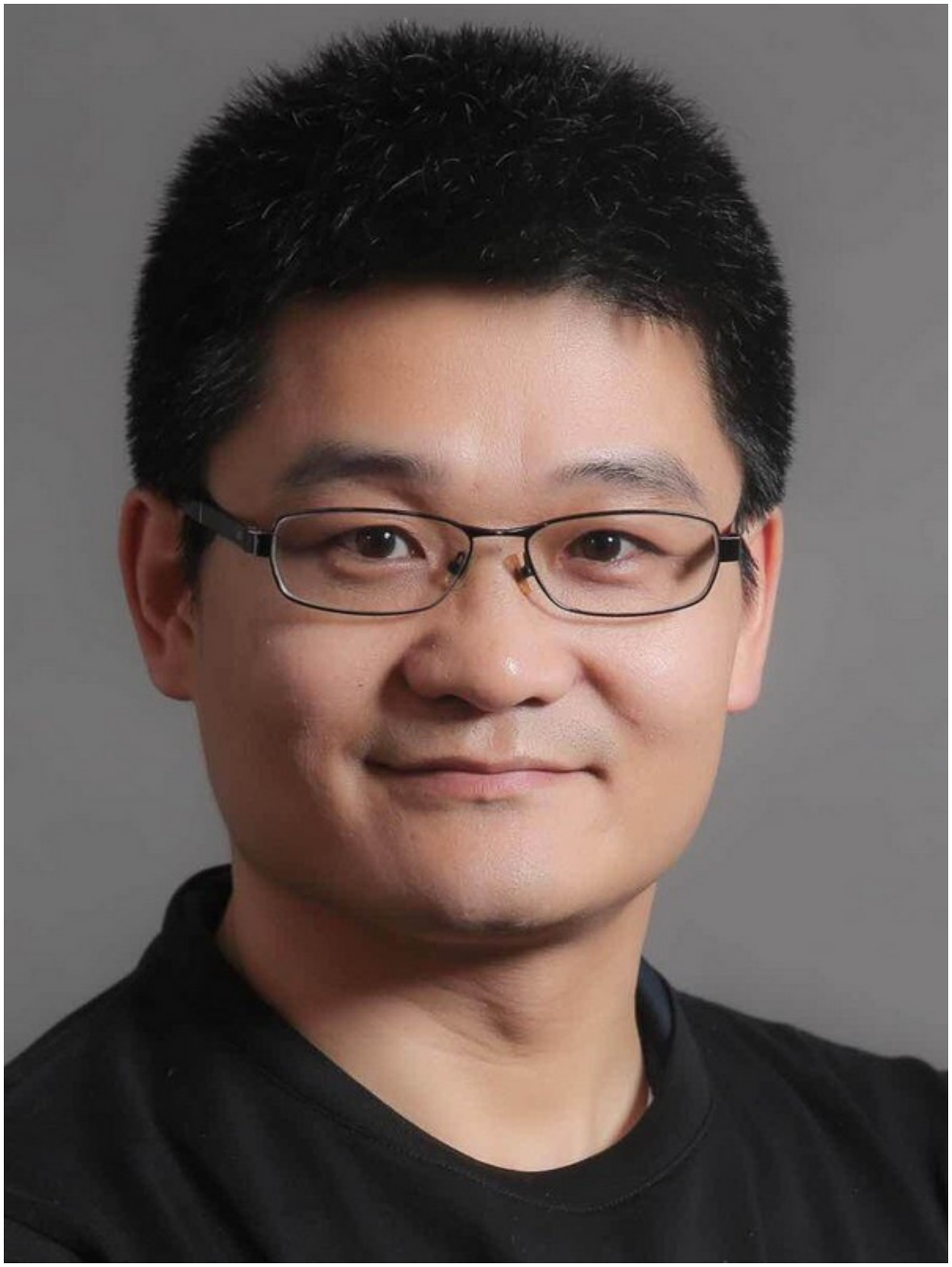}}]{Shiguang Shan}
(M’04-SM’15) received Ph.D. degree in computer science from the Institute of Computing Technology (ICT), Chinese Academy of Sciences (CAS), Beijing, China, in 2004. He has been a full Professor of this institute since 2010 and now the deputy director of CAS Key Lab of Intelligent Information Processing. His research interests cover computer vision, pattern recognition, and machine learning. He has published more than 300 papers, with totally more than 20,000 Google scholar citations. He served as Area Chairs for many international conferences including CVPR, ICCV, AAAI, IJCAI, ACCV, ICPR, FG, etc. And he was/is Associate Editors of several journals including IEEE T-IP, Neurocomputing, CVIU, and PRL. He was a recipient of the China’s State Natural Science Award in 2015, and the China’s State S\&T Progress Award in 2005 for his research work.
\end{IEEEbiography}


\begin{IEEEbiography}
[{\includegraphics[width=1in,height=1.25in,clip,keepaspectratio]{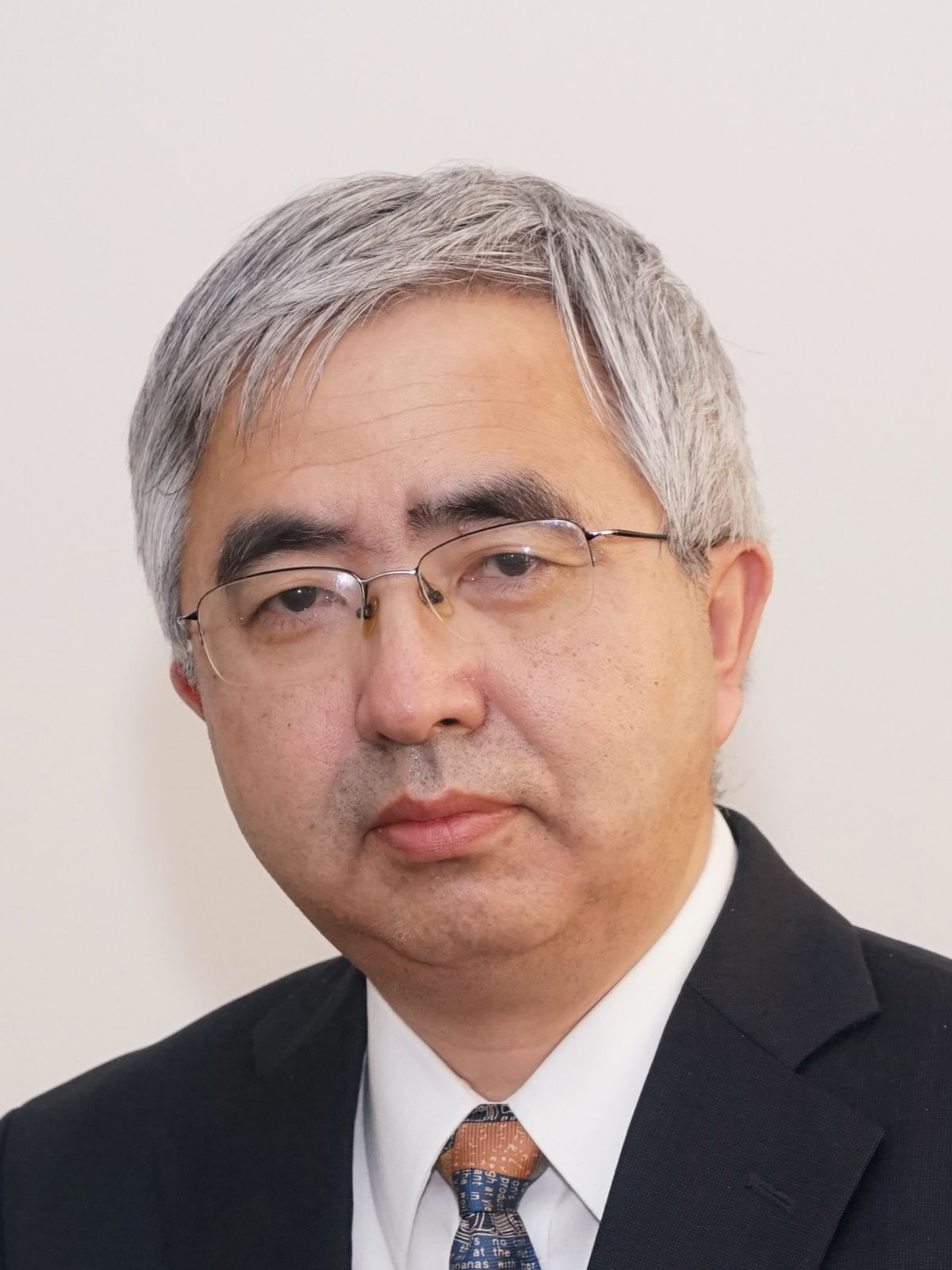}}]{Xilin Chen}
is a professor with the Institute of Computing Technology, Chinese Academy of Sciences (CAS). He has authored one book and more than 300 papers in refereed journals and proceedings in the areas of computer vision, pattern recognition, image processing, and multimodal interfaces. He is currently an information sciences editorial board member of Fundamental Research, an editorial board member of Research, a senior editor of the Journal of Visual Communication and Image Representation, and an associate editor-in-chief of the Chinese Journal of Computers, and Chinese Journal of Pattern Recognition and Artificial Intelligence. He served as an organizing committee member for multiple conferences, including general co-chair of FG13 / FG18, program co-chair of ICMI 2010. He is a fellow of the ACM, IEEE, IAPR, and CCF.
\end{IEEEbiography}


\end{document}